\documentclass{article}

\usepackage{arxiv}

\usepackage[utf8]{inputenc} % allow utf-8 input
\usepackage[T1]{fontenc}    % use 8-bit T1 fonts
\usepackage{hyperref}       % hyperlinks
\usepackage{url}            % simple URL typesetting
\usepackage{booktabs}       % professional-quality tables
\usepackage{amsfonts}       % blackboard math symbols
\usepackage{nicefrac}       % compact symbols for 1/2, etc.
\usepackage{microtype}      % microtypography
\usepackage{lipsum}
\usepackage{graphicx}
\usepackage{appendix}
\usepackage{amsmath}
\usepackage{multirow}
\usepackage{xspace}
\usepackage{hyperref}
\hypersetup{
    colorlinks=true,
    linkcolor=blue,
    filecolor=magenta,      
    urlcolor=cyan,
}

\author{
  Heejoon Koo\\
  University College London\\
  heejoon.koo.17@alumni.ucl.ac.uk
   \And
   To Eun Kim\\
   Carnegie Mellon University\\
   toeunkim@cmu.edu
}

\title{A Comprehensive Survey on \protect\\ Generative Diffusion Models for Structured Data}

\begin{document}
\maketitle

\begin{abstract}

In recent years, generative diffusion models have achieved a rapid paradigm shift in deep generative models by showing groundbreaking performance across various applications. Meanwhile, structured data, encompassing tabular and time series data, has been received comparatively limited attention from the deep learning research community, despite its omnipresence and extensive applications. Thus, there is still a lack of literature and its reviews on structured data modelling via diffusion models, compared to other data modalities such as visual and textual data. To address this gap, we present a comprehensive review of recently proposed diffusion models in the field of structured data. First, this survey provides a concise overview of the score-based diffusion model theory, subsequently proceeding to the technical descriptions of the majority of pioneering works that used structured data in both data-driven general tasks and domain-specific applications. Thereafter, we analyse and discuss the limitations and challenges shown in existing works and suggest potential research directions. We hope this review serves as a catalyst for the research community, promoting developments in generative diffusion models for structured data.

\end{abstract}

\keywords{Survey \and Diffusion Models \and Generative Models \and Structured Data \and Tabular Data \and Time Series Data}

\section{Introduction}

% An explanation on structured data
Structured data is characterised as its standardised format and ubiquitous across various domains. This type of data can be divided into two categories. The first is tabular data, where information is arranged into rows and columns, representing individual records and their respective attributes. The second category is time series data, which is sequential observations obtained at successive time intervals. The structured data has been extensively applied to many tasks, such as financial modelling \cite{lin2021forecasting}, fraud detection \cite{tiwari2021credit}, click-through rate (CTR) prediction \cite{zhou2019deep}, clinical event prediction \cite{beunza2019comparison}, counterfactual estimation \cite{liu2022practical}, and so forth. Enhancing predictive performance and robustness in these applications can provide significant benefits for both end users and organisations offering such solutions. Thus, structured data modelling has been a long-standing research topic in both academia and industry. 

% A method shift from traditional machine learning to deep generative modeling in structured data
Over the past decade, deep learning has revolutionised numerous fields, including computer vision and language modelling \cite{Kim2021AGCNAG, song2021solving, anuchitanukul2022surf, kim2022multi, Kim_2022_CVPR, gu2022vector, li2022diffusion, condita2022alexaprize, yu2022latent, chung2023diffusion}. This substantial advancement can be largely attributed to data-driven deep learning technologies, which have received considerable focus from the research community \cite{lecun2015deep}. Nevertheless, traditional machine learning techniques are still widely used for structured data and the volume of literature on structured data via deep learning is relatively insufficient. This is primarily due to the challenges that deep learning methodologies face when applied to structured data. First, datasets related to structured data are generally smaller than those for visual or textual data, thus limiting the full exploitation of deep learning's expressiveness. As a result, traditional machine learning methods are still being broadly utilised \cite{elsayed2021we, shwartz2022tabular}. Moreover, dataset complications, such as mixed type of data (both continuous and categorical types), the absence of correlation amongst columns and rows, and the necessity for domain knowledge-guided feature engineering, have made the application of deep learning to structured data a complex task \cite{kadra2021well}. Kadra \textit{et al}. \cite{kadra2021well} thus referred to structured data as the final \textit{unconquered castle} in deep learning research community. However, structured data modelling via deep generative models including variational auto-encoders (VAE) \cite{kingma2013auto} and generative adversarial networks (GAN) \cite{goodfellow2020generative} has been continually explored for applications such as data synthesis against privacy concerns \cite{vardhan2020generating, esteban2017real}, scenario-based simulations \cite{wiese2020quant}, and imputations \cite{yoon2018gain}. Additionally, such generative modelling has also been explored to improve predictive performance on structured data. 

% An explanation on the diffusion models
Score-based diffusion models \cite{song2019generative, ho2020denoising, song2020score} recently have become very prominent in various domains and applications, due to their superior capabilities compared to previous deep generative model families. Diffusion models were initially introduced by Sohl-Dickstein \textit{et al}. \cite{sohl2015deep}, inspired by non-equilibrium statistical physics. Subsequently, they are further developed by \cite{song2019generative, ho2020denoising}, verifying their potential in comparison with other state-of-the-art generative models in image synthesis tasks. Hereafter, they have shown exceptional performance across a variety of challenging tasks in different domains, \textit{e.g.,} inverse problems \cite{song2021solving, chung2023diffusion}, text driven image synthesis and editing \cite{Kim_2022_CVPR, gu2022vector}, language modelling \cite{li2022diffusion, yu2022latent}, 3D molecule generation \cite{huang2022mdm, xu2023geometric}. However, albeit its growing body of research, the attention on structured data modelling through diffusion models still remains insufficient. With the aims of promoting future research, we provide a holistic overview of the generative diffusion models for structured data in this paper.

% A description of remaining structure of this paper
The remaining structure of this paper is outlined as follows. As preliminaries, a brief introduction on backgrounds of generative diffusion models is described in Section \ref{sec:backgrounds}. Then, this survey dichotomises the existing works into two main categories: data-driven general tasks (Section \ref{sec:general}) and domain-specific applications (Section \ref{sec:domain}). Each section provides an overview and describes cutting-edge works along with their key technical novelties. In Section \ref{sec:challengesnopportunities}, the limitations and challenges shown in existing works are discussed with potential research directions. Finally, we conclude the survey in Section \ref{sec:conclusion}. 

\textbf{What Sets Our Survey Apart from Others} There are several existing surveys on diffusion models, either covering algorithmic developments, various data modalities and applications \cite{yang2022diffusion, cao2022survey}, or focusing on specific data modalities such as vision \cite{croitoru2023diffusion}, language \cite{zhu2023diffusion}, and graphs \cite{zhang2023survey}, or concentrating on medical imaging \cite{kazerouni2023diffusion} in a domain specific manner. Lin \textit{et al}. \cite{lin2023diffusion} is covering structured data, but their work is limited to time series applications. Recognising the gap in the literature review concerning diffusion models for the structured data, this work aims to address this deficiency by presenting the \textit{first} comprehensive survey dealing with structured data, including both tabular and time series data, and their related applications.

\begin{figure}[t]
\begin{center}
\includegraphics[width=0.8\linewidth]{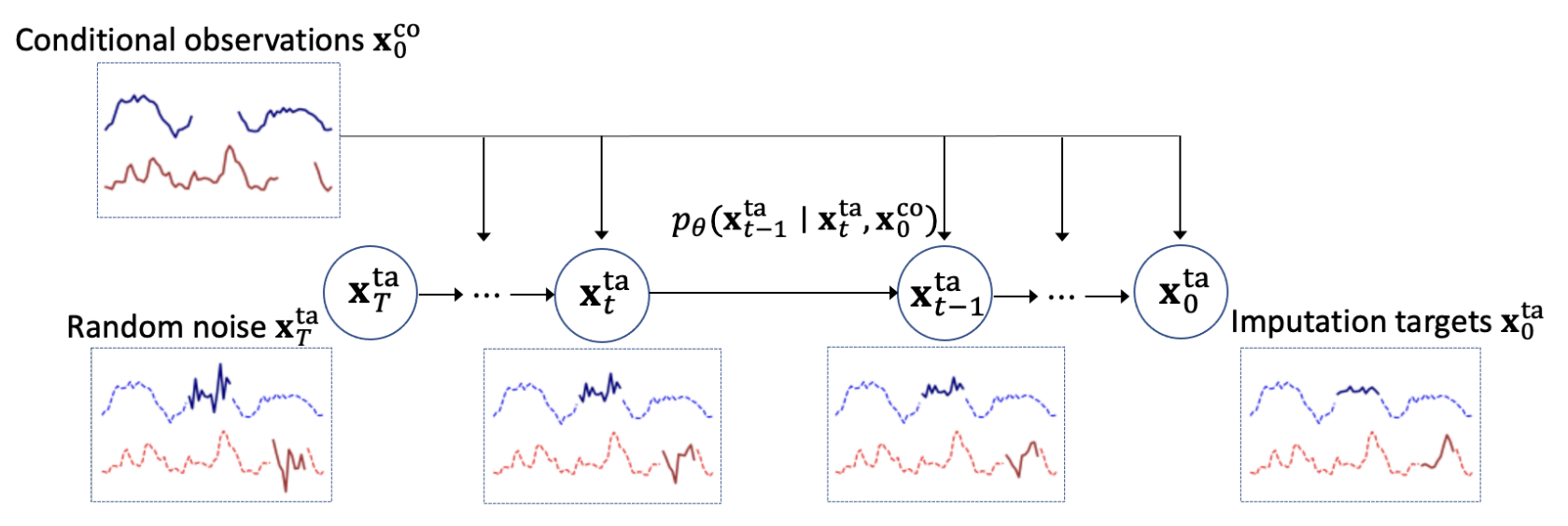}
\end{center}
\caption{A Process of Time Series Imputation via Diffusion Models. The reverse process $ p_\theta $ progressively removes random noise to generate plausible time series data, conditioned on observed values $ \mathbf{x}^\mathrm{co}_0 $. The dashed lines represent observed values, while the solid lines are targets for imputation, denoted as $ \mathbf{x}^\mathrm{ta}_0 $, where $ t $ corresponds to a specific diffusion step out of the total step $ T $. This illustration is derived from  CSDI \cite{tashiro2021csdi} to provide an intuitive understanding of time series modelling through diffusion models.}
\label{fig:timeseriesimputation}
\end{figure}

\section{Backgrounds on Score-based Diffusion Models}
\label{sec:backgrounds}

% Backgrounds on Diffusion Models 
Score-based diffusion models are a class of probabilistic generative models that learn to reversal of the data destruction processes that gradually injects noise, to yield high-quality and realistic synthesised data samples. In other words, the training procedure involves two steps: the forward diffusion process and the subsequent backward denoising process. Despite their diverse applications across various data modalities, the design of the forward and backward processes categorises the current research into three main frameworks: denoising diffusion probabilistic models (DDPMs) \cite{ho2020denoising, sohl2015deep}, score-based models (SGMs) \cite{song2019generative}, and stochastic differential equations (SDEs) \cite{song2020score}. Therefore, this section provides a concise review on three subcategories of the diffusion models with mathematical formulae. We present only the essential derivations, thus we recommend referring to the original papers for comprehensive equations.

\subsection{Denoising Diffusion Probabilistic Models (DDPMs)}
\label{sec:backgrounds_ddpms}

% An overview on ddpm
DDPMs design the forward and reverse processes via dual Markov chains \cite{ho2020denoising, sohl2015deep}. The forward process involves the diffusion of data with pre-determined noise, such as Gaussian noise, whilst the reverse process employs deep neural networks to sequentially eliminate noise and recovers the original data. 

% An overview on forward process of ddpm
\textbf{Forward Pass} Suppose that there is a clean data point $\mathbf{x_0}$ drawn from a data distribution $q(\mathbf{x_0})$. Then, the forward diffusion process progressively perturbs the clean original data distribution by adding Gaussian noise, ultimately converging towards the standard Gaussian distribution $\mathbf{z}_T$. During the diffusion step $T$, noised latent variables $\mathbf{x}_1$, $\mathbf{x}_2$, ... , $\mathbf{x}_{T}$ are yielded. In other words, it generates $\mathbf{x}_{T}$ using sequential transition kernel $q(\mathbf{x}_{t} | \mathbf{x}_{t-1})$, which can be formulated as below: 

\begin{equation} \label{eq:ddpm_forward1}
    q(\mathbf{x}_t | \mathbf{x}_{t-1}) := \mathcal{N}(\mathbf{x}_{t} ; \sqrt{1-\beta_{t}} \mathbf{x}_{t-1}, \beta_{t} \mathbf{I}), \hspace*{1mm} \forall {t} \in \{ 1, \cdots, T \}
\end{equation}

thus, the forward process is defined by using a series of transition kernels:

\begin{equation} \label{eq:ddpm_forward2}
    q(\mathbf{x}_{1:T} | \mathbf{x}_{0}) := \prod_{t=1}^T {q}(\mathbf{x}_{t} | \mathbf{x}_{t-1})
\end{equation}

where $ \beta_{T} \in (0, 1) $ is a variance schedule that controls step sizes, and $ \mathbf{I} $ is the identity matrix with the same dimension as input data $ \mathbf{x_0} $. Also, $ \mathcal{N}(\mathbf{x} ; \mu, \sigma) $ is the Gaussian distribution of $ \mathbf{x} $ with the mean $ \mu $ and covariance $ \sigma $. Let $ \alpha_{t} := 1 - \beta_{t} $ and $ \bar{\alpha_{t}} := \prod_{s=1}^{t} \alpha_{s} $ and $ \epsilon $ as a Gaussian noise, it becomes feasible to yield a noisy sample from an arbitrary step from the distribution, conditioned on the original input $ \mathbf{x_0} $ \cite{vignac2022digress} as follows:

\begin{equation} \label{eq:ddpm_forward3}
    q(\mathbf{x}_{t} | \mathbf{x}_{0}) := \mathcal{N}(\mathbf{x}_{t} ; \sqrt{\bar{\alpha_{t}}} \mathbf{x}_0, \sqrt{1 - \bar{\alpha_{t}}} \mathbf{I}),
\end{equation}

\begin{equation} \label{eq:ddpm_forward4}
    \mathrm{where} \hspace*{1mm} \mathbf{x}_{t} = \sqrt{\bar{\alpha_{t}}}  \mathbf{x}_0 + \sqrt{1 - \bar{\alpha_{t}}} \epsilon.
\end{equation}

% An overview on reverse process of ddpm
\textbf{Reverse Process} With the forward process, the reverse process removes the noise at each step in reverse time direction until the destroyed original data is reconstructed. We start with $ p_\theta(\mathbf{x}_{T}) $ to generate $ p_\theta(\mathbf{x}_0) $ that obeys the true data distribution $ {q}(\mathbf{x}_0) $. Again, $ \mathbf{x}_{T} \sim \mathcal{N}(\mathbf{0}, \mathbf{I}) $. A series of reverse Gaussian transition kernels $ {p}_\theta $, where $\theta $ denotes the learnable parameters, is parameterised in the form of a deep neural network as:
 
\begin{equation} \label{eq:ddpm_backward1}
    {p}_\theta (\mathbf{x}_{0:T}) := {p}(\mathbf{x}_{T}) \prod_{t=1}^T {p}_\theta(\mathbf{x}_{t-1} | \mathbf{x}_{t}),
\end{equation}

\begin{equation} \label{eq:ddpm_backward2}
    {p}_\theta (\mathbf{x}_{t-1} | \mathbf{x}_{t}) := \mathcal{N}(\mathbf{x}_{t-1} ; \mu_\theta(\mathbf{x}_{t}, {t}), \sigma_\theta(\mathbf{x}_{t}, {t}) \mathbf{I}).
\end{equation}

Additionally, the model parameterises both the mean $ \mu_\theta(\mathbf{x}_{t}, {t})$ and variance $\sigma_\theta(\mathbf{x}_{t}, {t})$. The model learns to approximate the true data distribution during the reverse process, which is achieved through optimising variational upper bound on negative log-likelihood (NLL):

\begin{equation} \label{eq:ddpm_vlb}
        \mathbb{E}\left[-\log p_\theta\left(\mathbf{x}_0\right)\right] \leq \mathbb{E}_q\left[-\log \frac{p_\theta\left(\mathbf{x}_{0: T}\right)}{q\left(\mathbf{x}_{1: T} \mid \mathbf{x}_0\right)}\right] \\
        =\mathbb{E}_q\left[-\log p\left(\mathbf{x}_T\right)-\sum_{t \geq 1} \log \frac{p_\theta\left(\mathbf{x}_{t-1} \mid \mathbf{x}_t\right)}{q\left(\mathbf{x}_t \mid \mathbf{x}_{t-1}\right)}\right] \\
        :=-L_{\mathrm{VLB}}.
\end{equation}

It can be rewritten using Kullback–Leibler divergence (KL divergence) as:

\begin{equation} \label{eq:ddpm_kl}
    \mathbb{E}_q \bigg[ \underbrace{D_\mathrm{KL}{(q(\mathbf{x}_T|\mathbf{x}_0)}{p(\mathbf{x}_T))}}_{L_T} + \sum_{t > 1} \underbrace{D_\mathrm{KL}{(q(\mathbf{x}_{t-1}|\mathbf{x}_t,\mathbf{x}_0)}  \hspace*{1mm} || \hspace*{1mm} {p_\theta(\mathbf{x}_{t-1}|\mathbf{x}_t))}}_{L_{t-1}} \underbrace{-\log p_\theta(\mathbf{x}_0|\mathbf{x}_1)}_{L_0} \bigg].
\end{equation}

The equation \ref{eq:ddpm_kl} is decomposed into three parts: $ L_T $, $ L_{t-1} $, and $ L_0 $. First, $ L_T $ minimises the KL divergence between $ q(\mathbf{x}_T|\mathbf{x}_0) $ and a standard Gaussian distribution $ p(\mathbf{x}_T) $. Next, in $ L_{t-1} $, the $p_\theta(\mathbf{x}_{t-1}|\mathbf{x}_t)$ can be computed against posteriors from the forward process. The last term $ L_0 $ denotes a negative log-likelihood. Especially, we condition $ L_{t-1} $ on $\mathbf{x}_0$ to make it tractable:

\begin{equation} \label{eq:ddpm_conditioning_1}
    q(\mathbf{x}_{t-1}|\mathbf{x}_{t}, \mathbf{x}_0) =  \mathcal{N}(\mathbf{x}_{t-1}; \tilde{\mu}_t(\mathbf{x}_{t}, \mathbf{x}_0), \tilde{\beta_{t}} \mathbf{I})
\end{equation}

\begin{equation} \label{eq:ddpm_conditioning_2}
    \mathrm{where} \hspace*{1mm} \tilde\mu_t(\mathbf{x}_t, \mathbf{x}_0) := \frac{\sqrt{\bar\alpha_{t-1}}\beta_t }{1-\bar\alpha_t}\mathbf{x}_0 + \frac{\sqrt{\alpha_t}(1- \bar\alpha_{t-1})}{1-\bar\alpha_t} \mathbf{x}_t \hspace*{1mm} \mathrm{and} \hspace*{1mm} \tilde\beta_t := \frac{1-\bar\alpha_{t-1}}{1-\bar\alpha_t}\beta_t.
\end{equation}

Considering the equation \ref{eq:ddpm_backward2}, \ref{eq:ddpm_conditioning_1} and \ref{eq:ddpm_conditioning_2}, the term $ L_{t-1} $ in equation \ref{eq:ddpm_kl} can be reformulated as:

\begin{equation} \label{eq:ddpm_lt1}
  L_{t-1} = \mathbb{E}_{q} \bigg[ { \frac{1}{2\sigma_t^2} \|\tilde \mu_t(\mathbf{x}_t, \mathbf{x}_0) - \mu_\theta(\mathbf{x}_t, t)\|^2 } \bigg] + C
\end{equation} 

% Reparameterisation by Ho et al.
where $ C $ does not depend on $ \theta $, thus considered as a constant. Especially, Ho \textit{\textit{et al}}. \cite{ho2020denoising} highlight that instead of parameterising the mean $ \mu_\theta(\mathbf{x}_{t}, {t})$, predicting noise vector $ \epsilon $ at each time step $ {t} $ in the forward process by parameterising $ \epsilon_\theta(\mathbf{x}_t, t) $ enhances training efficiency and improves sample quality. They rewrite the equation \ref{eq:ddpm_lt1} as below:

\begin{equation} \label{eq:ddpm_ho}
    \mathbb{E}_{t \sim \mathcal{U} (1,T), \mathbf x_0 \sim q(\mathbf x_0), \epsilon \sim \mathcal{N}(\mathbf{0},\mathbf{I})} \bigg[ {\lambda(t)  \left\| \epsilon - \epsilon_\theta(\mathbf{x}_t, t) \right\|^2} \bigg]
\end{equation}

where $ \mathcal{U} (1,T) $ denotes a uniform distribution, $ \lambda(t) $ is a weighting function to adjust the scales of noise evenly, and $ \epsilon_\theta $ is a deep parameterised model that predicts the Gaussian noise $ \epsilon \sim \mathcal{N}(\mathbf{0}, \mathbf{I}) $. Once trained, it samples $ \mathbf{x}_0 $ that resembles the original data. 

% Introduction of multinomial diffusion
Whilst the DDPM can only be applied to continuous data, such as image and audio, its application can be extended beyond these types. Hoogeboom \textit{et al}. \cite{hoogeboom2021argmax} have introduced a multinomial diffusion, specifically designed for categorical data. This model corrupts the data using a discrete Markovian process, then restores the original data through a reversal process. It is particularly crucial in certain data modalities and applications in which the usage of categorical data types is involved: tables, text, segmentation maps, and such \cite{zhu2023diffusion, hoogeboom2021argmax, kotelnikov2022tabddpm}. We provide an explanation of multinomial diffusion in the Appendix \ref{appendix:multinomialdiffusion}.

\subsection{Score-based Generative Models (SGMs)}
\label{sec:backgrounds_sgms}

% An overview of SGMs
Here, we employ a score function of a probabilistic density function, denoted as $ {p(\mathbf{x})} $. The score function, $ \nabla_{\mathbf{x}} \log {p(\mathbf{x})} $, is defined as the gradient of the logarithm of the probabilistic density with respect to the input $ \mathbf{x} $. In order to estimate the score function, we train a deep neural network $ {s_\theta(\mathbf{x})} $, where $ \theta $ represents the learnable parameters and $ \mathbf{x} $ is the input data, to approximate the score of the original data distribution $p(\mathbf{x})$. It is mathematically formulated as following:

\begin{equation} \label{eq:sgms1}
    \mathbb{E}_{\mathbf{x} \sim {p(\mathbf{x})}} || {s_\theta(\mathbf{x})} - \nabla_{\mathbf{x}} \log {p(\mathbf{x})} ||_2^2.
\end{equation}

% Introduction to works on score matching 
Nevertheless, it is computationally infeasible to obtain $ \nabla_{\mathbf{x}} \log {p(\mathbf{x})} $ in the context of high-dimensional data and deep neural networks. Thus, there are various works to address the issue: score matching \cite{hyvarinen05a}, denoising score matching \cite{song2019generative, vincent2011connection}, and sliced score matching \cite{song2020sliced}. 

% An explanation on NCSN
In particular, noise-conditioned score network (NCSN) (a.k.a score matching with langevin dynamics, SMLD) \cite{song2019generative} emphasise issues regarding manifold hypothesis. In circumstances where real-world data concentrate on the low-dimensional manifolds that are embedded within a high-dimensional space, the estimated score functions are inevitably imprecise in regions of low density. Thus, they propose to inject the random Gaussian noise to the original data with a sequence of intensifying scale, making the data distribution more amenable to SGMs, and estimate the score corresponding to each noise level. Mathematically, we have a sequence of Gaussian noise scales $ 0 < \sigma_1 < \sigma_2 < \cdots < \sigma_{t} < \cdots < \sigma_{T} $, thus $ p_{\sigma_1}(\mathbf{x}) \approx p(\mathbf{x}_0) $, $ p_{\sigma_T}(\mathbf{x}) \approx \mathcal{N}(\mathbf{x} ; 0, \mathbf{I}) $ and $ p_{\sigma_t}(\mathbf{x}_t | \mathbf{x}) \approx \mathcal{N}(\mathbf{x}_t ; \mathbf{x}, \sigma_{t}^2 \mathbf{I}) $. Moreover, a single noise-conditioned score networks $ s_\theta(\mathbf{x}, \sigma_t) $ aims to approximate the gradient logarithm density function $ \nabla_\mathbf{x} \log p_{\sigma_{{t}}}(\mathbf{x}) $. For a specific $ \mathbf{x}_{t} $, the derivation of $ \nabla_\mathbf{x} \log p_{\sigma_{t}}(\mathbf{x}) $ is:

\begin{equation} \label{eq:sgms2}
    \nabla_\mathbf{x} \log p_{\sigma_{t}}(\mathbf{x}_t | \mathbf{x}) = - \frac{\mathbf{x}_t - \mathbf{x}}{\sigma_t}.
\end{equation}

In particular, the directions of the gradients are towards regions where the density of samples is high. Then, the combination of the equation \ref{eq:sgms1} and \ref{eq:sgms2} with a weighting function $ \lambda\left(\sigma_t\right) $ derives a new equation as:

\begin{equation} \label{eq:sgms3}
	\frac{1}{T} \sum_{t=1}^T \lambda\left(\sigma_t\right) \mathbb{E}_{p(\mathbf{x})} \mathbb{E}_{\mathbf{x}_t \sim p_{\sigma_t}\left(\mathbf{x}_t \mid \mathbf{x}\right)}\left\|s_\theta\left(\mathbf{x}_t, \sigma_t\right)+\frac{\mathbf{x}_t-\mathbf{x}}{\sigma_t}\right\|_2^2.
\end{equation}

With the annealed Langevin dynamics, new samples are generated by a progressive denoising process from the prior Gaussian distribution. The annealed Langevin dynamics exploits a Markov chain Monte Carlo (MCMC) to draw a sample from the distribution $ p(\mathbf{x}) $ using the score function $ \nabla_\mathbf{x} \log p(\mathbf{x}) $. It recursively samples $ \mathbf{x}_i $ as follows:

\begin{equation} \label{eq:Langevin}
    \mathbf{x}_i=\mathbf{x}_{i-1}+\frac{\gamma}{2} \nabla_\mathbf{x} \log p(\mathbf{x})+\sqrt{\gamma} \omega_i, \forall i \in \{ 1, \cdots, N\}
\end{equation}

where $ \omega_i \sim \mathcal{N}(\mathbf{0}, \mathbf{I}) $, $ \gamma $ determines both magnitude and direction of the score update. After executing $ N $ iterations, the $ \mathbf{x} $ becomes a sample derived from the original distribution  $ p(\mathbf{x}) $. Notably, in NCSN \cite{song2019generative}, the magnitude of $ \omega_i $ undergoes gradual decrement, thereby subtly introducing uncertainty and preventing the model from mode failure. 

\subsection{Stochastic Differential Equations (SDEs)}
\label{sec:backgrounds_sdes} 

% An overview on SDEs
Since the objective forms of both SGMs \cite{song2019generative} and DDPMs \cite{ho2020denoising} are similar, Song \textit{et al}. \cite{song2020score} have integrated and further generalised these into a single framework where the number of noise scales is extended to infinity via stochastic differential equations (SDEs). The corresponding continuous diffusion process can be described using Itô SDE as:

\begin{equation} \label{eq:sdeforward}
	\mathrm{d} \mathbf{x}=\mathbf{f}(\mathbf{x}, t) \mathrm{d} t+g(t) \mathrm{d} \mathbf{w}, t \in [0, T].
\end{equation}

where $ \mathbf{f} $ is a drift coefficient of $ 
\mathbf{x}(t) $ and $ \mathbf{g} $ is a diffusion coefficient that is interwined with standard Wiener process $ \mathbf{w} $. Also, $ \mathrm{d}t $ is a infinitesimal negative time step. 

Similar to DDPMs and SGMs, $ \mathbf{x_0} $ and $ \mathbf{x}_T $ denote data samples from the clean distribution $ p_0 $ and standard Gaussian distribution  $ p_T $, respectively. Accordingly, it synthesises new samples from the known prior distribution $ p_T $, by solving the reverse-time SDE:

\begin{equation} \label{eq:sde_backward}
	\mathrm{d} \mathbf{x}=\left[\mathbf{f}(\mathbf{x}, t)-g^2(t) \nabla_\mathbf{x} \log p_t(\mathbf{x}) \right] \mathrm{d} t+g(t) \mathrm{d} \bar{\mathbf{w}}
\end{equation}

where $ \bar{\mathbf{w}} $ is the reverse standard Brownian motion. The solution to reverse-time SDE is to approximate a time-dependent deep neural network $ s_\theta(\mathbf{x}, t) $ to a score function $ \nabla_\mathbf{x} \log p_t{(\mathbf{x})} $ via score matching objective function. Instead of directly approximating the score function which is computationally intractable, it estimates the transition probability $ \nabla_{\mathbf{x}_{t}}{\log{{p}_{t}\left( \mathbf{x}_{t} \middle| \mathbf{x}_{0} \right)}} $ that follows the Gaussian distribution during the forward diffusion process as:

\begin{equation} \label{eq:sde_objective}
    \mathbb{E}_{t,\mathbf{x}_{0},\mathbf{x}_{t}}\left\lbrack \lambda(t)\left| \left| {s_{\theta}\left( \mathbf{x}_{t},t \right) - \nabla_{\mathbf{x}_{t}}{\log{{p}_{0t}\left( \mathbf{x}_{t} \middle| \mathbf{x}_{0} \right)}}} \right| \right|^{2}_{2} \right\rbrack.
\end{equation}

Here, $ p_{0t}\left( \mathbf{x}_{t} \middle| \mathbf{x}_{0} \right) $ denotes the transition kernel of $ \mathbf{x}_{t} $ given $ \mathbf{x}_{0} $ and $ \lambda(t) $ is the weighting function. Upon completion of the training process, we can generate samples employing various techniques such as the Euler-Maruyama (EM), Prediction-Correction (PC), or Probability Flow ODE method.

% An explanation on reverse-SDE solvers
EM solves the equation \ref{eq:sde_backward} by using a simple discretisation technique, where $ \mathrm{d} \mathbf{x} $ is substituted with $ \triangle \mathbf{t} $ and $ \mathrm{d} \bar{\mathbf{w}} $ is replaced by the Gaussian noise $ z \sim \mathcal{N}(\mathbf{0}, \Delta t \mathbf{I}) $. In PC method, it operates in a sequential manner, alternating between predictor and corrector. The predictor can employ any numerical solver for the reverse-time SDE following a fixed discretisation strategy, such as the EM method. Subsequently, the corrector can be any score-based Markov chain Monte Carlo (MCMC) method, like annealed Langevin dynamics. Thus, the equation \ref{eq:Langevin} can be solved using Langevin dynamics. In Probability Flow ODE method, the equation \ref{eq:sdeforward} can be reformulated into an ODE given by:

\begin{equation} \label{eq:sde_ode}
    \mathrm{d} \mathbf{x}=\left[\mathbf{f}(\mathbf{x}, t)-\frac{1}{2} g^2(t) \nabla_\mathbf{x} \log p_t(\mathbf{x})\right] \mathrm{d}t.
\end{equation}

This equation maintains the identical marginal probability density $ p_t(t) $, as that of the SDE. Thus, sampling via solving the above reverse-time ODE is equivalent to solving the time reversal SDE.

\section{Data-driven General Tasks}
\label{sec:general}

% An overview on Section 3 and 4
In this and subsequent sections, we delve into an in-depth review of diffusion-based methodologies for structured data, which are divided into two main categories: data-driven general tasks and domain-specific applications. For data-driven tasks, we focus on two data modalities: tabular and time series data. When it comes to tabular data modelling, we focus on generation and imputation, whilst for time series data, we delve into generation, imputation, and forecasting. Moving on to the domain-specific applications, we divide them into three categories: electronic health records (EHR), bioelectrical signal processing, and recommendation systems (RecSys). For EHR and bioelectrical signal processing, we explore both generation and specific tasks, with forecasting for EHR and enhancement for bioelectrical signal processing. Table \ref{table:summary} presents a quick summary on these categories along with the corresponding works. For further details, such as information on the generative modelling framework, specific datasets used in the experiments and links to accessible code, please refer to the Appendix \ref{appendix:comprehensive}.

\begin{table}[ht]
\caption{A hierarchical table on generative diffusion models for structured data}
\centering
\begin{tabular}{c|cc|c}
\hline
\textbf{Categories} & \multicolumn{1}{c|}{\textbf{Data Type}} & \textbf{Task} & \textbf{Papers} \\ \hline
\multirow{14}{*}{Data-driven General Task} & \multicolumn{1}{c|}{\multirow{5}{*}{Tabular Data}} & \multirow{4}{*}{Generation} & TabDDPM \cite{kotelnikov2022tabddpm} \\ \cline{4-4} 
 & \multicolumn{1}{c|}{} &  & SOS \cite{kim2022sos} \\ \cline{4-4} 
 & \multicolumn{1}{c|}{} &  & STaSy \cite{kim2022stasy} \\ \cline{4-4} 
 & \multicolumn{1}{c|}{} &  & CoDi \cite{lee2023codi} \\ \cline{3-4} 
 & \multicolumn{1}{c|}{} & Imputation & TabCSDI \cite{zheng2022diffusion} \\ \cline{2-4} 
 & \multicolumn{1}{c|}{\multirow{9}{*}{Time Series Data}} & Generation & TSGM \cite{lim2023regular} \\ \cline{3-4} 
 & \multicolumn{1}{c|}{} & \multirow{3}{*}{Imputation} & CSDI \cite{tashiro2021csdi} \\ \cline{4-4} 
 & \multicolumn{1}{c|}{} &  & SSSD$^\mathrm{S4}$ \cite{alcaraz2022diffusion} \\ \cline{4-4} 
 & \multicolumn{1}{c|}{} &  & DSPD/CSPD \cite{bilovs2022modeling} \\ \cline{3-4} 
 & \multicolumn{1}{c|}{} & \multirow{5}{*}{Forecasting} & TimeGrad \cite{rasul2021autoregressive} \\ \cline{4-4} 
 & \multicolumn{1}{c|}{} &  & ScoreGrad \cite{yan2021scoregrad} \\ \cline{4-4} 
 & \multicolumn{1}{c|}{} &  & SSSD$^\mathrm{S4}$ \cite{alcaraz2022diffusion} \\ \cline{4-4} 
 & \multicolumn{1}{c|}{} &  & D$^{3}$VAE \cite{li2022generative} \\ \cline{4-4} 
 & \multicolumn{1}{c|}{} &  & DSPD/CSPD \cite{bilovs2022modeling} \\ \hline
\multirow{13}{*}{Domain-specific Applications} & \multicolumn{1}{c|}{\multirow{5}{*}{Electronic Health Records}} & \multirow{4}{*}{Generation} & EHRDiff \cite{yuan2023ehrdiff} \\ \cline{4-4} 
 & \multicolumn{1}{c|}{} &  & TabDDPM \cite{ceritli2023synthesizing} \\ \cline{4-4} 
 & \multicolumn{1}{c|}{} &  & MedDiff \cite{he2023meddiff} \\ \cline{4-4} 
 & \multicolumn{1}{c|}{} &  & EHR-DPM \cite{kuo2023synthetic} \\ \cline{3-4} 
 & \multicolumn{1}{c|}{} & Forecasting & TDSTF \cite{chang2023tdstf} \\ \cline{2-4} 
 & \multicolumn{1}{c|}{\multirow{3}{*}{Bioelectrical Signals}} & Generation & SSSD-ECG \cite{alcaraz2023diffusion} \\ \cline{3-4} 
 & \multicolumn{1}{c|}{} & \multirow{2}{*}{Enhancement} & DeScoD-ECG \cite{li2023descod} \\ \cline{4-4} 
 & \multicolumn{1}{c|}{} &  & DS-DDPM \cite{duan2023domain} \\ \cline{2-4} 
 & \multicolumn{2}{c|}{\multirow{5}{*}{Recommendation Systems}} & CODIGEM \cite{walker2022recommendation} \\ \cline{4-4} 
 & \multicolumn{2}{c|}{} & DiffuRec \cite{li2023diffurec} \\ \cline{4-4} 
 & \multicolumn{2}{c|}{} & DiffRec (Du \textit{et al}.) \cite{du2023sequential} \\ \cline{4-4} 
 & \multicolumn{2}{c|}{} & DiffRec (Wang \textit{et al}.) \cite{wang2023diffusion} \\ \cline{4-4} 
 & \multicolumn{2}{c|}{} & CDDRec \cite{wang2023conditional} \\ \hline
\end{tabular}
\label{table:summary}
\end{table}

\subsection{Tabular Data}

% \textbf{Overview} 
% \textbf{Problem Formulation} 
% \textbf{Methodologies} 

% An overview on tabular data
Tabular data, an organised data format in a rectangular grid of rows and columns, is one of the most universal data types in real-world applications. Nevertheless, handling tabular datasets directly using deep learning has been obstructed by various challenges, such as potential privacy issues and missing information due to data storage or human errors \cite{kim2022stasy, zheng2022diffusion}. Thus, deep generative models (including VAE \cite{kingma2013auto} and GAN \cite{goodfellow2020generative}) have been investigated for both tabular data synthesis \cite{vardhan2020generating, xu2019modeling} or imputation \cite{yoon2018gain, camino2019improving}. To further improve the performance, researchers have started to use diffusion models, a new paradigm in generative models, for the generation and imputation of tabular data.

\subsubsection{Generation}

% An explanation on tabDDPM
TabDDPM \cite{kotelnikov2022tabddpm} is the first pioneering work in the field of tabular data synthesis. To tackle mixed-type characteristics of tabular data, it employs the Gaussian diffusion to model continuous features and multinomial diffusion to model categorical features \cite{hoogeboom2021argmax}. To pre-process the mixed-type of data, they convert the continuous features using min-max scaler whilst they convert categorical features through one-hot encoding and process each feature using a separate diffusion process. During post-processing, they apply reverse scaling when generating continuous variables. For categorical features, they use softmax function, followed by a rounding operator. These pre- and post-processing techniques are widely used by the most of the follow-up works. 

TabDDPM \cite{kotelnikov2022tabddpm} uses class-conditional model for classification datasets and inserts target values as an additional feature for regression datasets. A simple MLP architecture is optimised through a combination of mean squared error (MSE) and KL divergence, with each term tailored to continuous and categorical features. The equation is:

\begin{equation} \label{eq:tabddpm}
    L^{TabDDPM}_t = L_t^{mse} + \frac{\sum_{i \leq C} L^{i}_t}{C}
\end{equation}

where $ C $ is the number of categorical features, $ L_t^{mse} $ is inherently equation \ref{eq:ddpm_ho}, and $ L_t^i $ minimises the KL divergence for each multinomial diffusion \cite{hoogeboom2021argmax}.
It outperforms other strong baselines in tabular data synthesis and even verified its efficacy on privacy criteria against SMOTE \cite{chawla2002smote}.

% An explanation on SOS
Next, to address the long-standing problem of class imbalance in tabular data, Kim \textit{et al}. \cite{kim2022sos} propose the first work on Score-based Over Sampling (SOS). It utilises style transfer to transform samples from the majority classes to the minority classes. In detail, the samples from the majority classes are corrupted by a forward SDE and recovered using the reverse SDE that originally learns to sample minority classes. Class-conditional fine-tuning is optionally applied to further improve the over-sampling performance. 

% An explanation on STaSy
STaSy \cite{kim2022stasy} is another seminal work on tabular data synthesis. Kim \textit{et al}. \cite{kim2022stasy} highlight that directly applying SDE \cite{song2020score} to the tabular data makes it challenging to learn the joint probability of multiple columns, particularly when there is little to no correlation amongst them. To mitigate the issue, they design architectures comprised of MLP residual blocks, which are dataset-dependent. They further integrate self-paced learning with denoising score matching objective to improve the performance. Fake tabular samples are generated by solving reverse SDEs using probability flow ODE method. It is also shown that sampling quality improves after fine-tuning in general.

% An explanation on CoDi
CoDi \cite{lee2023codi} addresses the training challenges that arise from mixed-data types by adopting a dual diffusion model approach; one model is for modelling continuous features, while the other is for discrete (categorical) features. These models are comprised of UNet-based architecture where convolutional layers are replaced with linear layers \cite{ronneberger2015u} and trained in a co-evolutionary fashion, being conditioned reciprocally. In other words, they receive each other's output as additional input. To elaborate, there is a pair of data, $ (\mathbf{x}_0^C, \mathbf{x}_0^D) $, and the perturbed data $ \mathbf{x}^C_t $ after $ t $ steps is a condition to $ \mathbf{x}^D_t $ in the discrete diffusion model, and vice versa. Then, the diffusion objective function for model handling continuous features is updated as follows:

\begin{equation} \label{eq:codi_1}
    L_{\mathrm{Diff_C}}(\theta_C) = \mathbb{E}_{t,\mathbf{x}_0^C,\boldsymbol{\epsilon}}\Big[\big\Vert\boldsymbol{\epsilon}-\boldsymbol{\epsilon}_{\theta_C}(\mathbf{x}_t^C, t \mid \mathbf{x}_t^D)\big\Vert^2\Big].
\end{equation}

For brevity, we only provide equations regarding the continuous features here. To further improve, triplet loss \cite{schroff2015facenet} is applied to each diffusion model independently. By learning a combination of two losses, it surpasses the other strong baselines in the real-world benchmark datasets.

\subsubsection{Imputation}

% An explanation on TabCSDI
TabCSDI \cite{zheng2022diffusion} is a novel approach on tabular data imputation built upon CSDI \cite{tashiro2021csdi}, which is for time-series data imputation and forecasting. To generate both continuous and categorical features of tabular data, it explores three conventional techniques: one-hot encoding, analog bits encoding \cite{chen2022analog}, and feature tokenisation \cite{gorishniy2021revisiting}. Especially, when utilising feature tokenisation, both continuous and categorical features undergo transformation via an embedding layer. Zheng \textit{et al}. \cite{zheng2022diffusion} emphasise the importance of feature tokenization as it mitigates the issue of column imbalance, which is commonly observed in the other two techniques. This approach leads to superior performance.

\subsection{Time Series Data}

% An overview on Time Series
Time series data consist of a sequence of observations recorded over regular intervals of time. It plays a crucial role in myriad fields, including finance, healthcare and climate. Its significance is particularly highlighted in decision-making processes, making time series forecasting methodologies evolve significantly from traditional statistical methods \cite{ho1998use}, to Recurrent Neural Networks (RNNs) \cite{hewamalage2021recurrent}, and Transformer-based models \cite{zhou2021informer}. 
However, the performance of predictive tasks using these models can degrade due to the presence of missing data, often attributable to device failures or human error \cite{cao2018brits}. As such, time series imputation strategies have also been explored \cite{ cao2018brits}. Concurrently, to facilitate scenario-based simulation and address privacy concerns, researchers have also focused on time series synthesis \cite{esteban2017real, wiese2020quant}. Aiming to comprehensively address the three core challenges—generation, imputation, and forecasting—researchers have delved into the time series modelling with generative diffusion models in recent years\footnote{To stay within our scope, we exclude time series modelling approaches based on spatio-temporal graph data, despite their relevance in certain applications. Our focus remains on the tabular and time series data modalities.}.
  
\subsubsection{Generation}
\label{sec:timeseriesgeneration}

% An explanation on TSGM
TSGM \cite{lim2023regular} is the first work on time-series synthesis. It generates fake time series data at each time step, based on the past synthesised observations. The framework follows a two-stage process: pre-training and main training, which are performed by encoder, decoder, and conditional score-matching network. During the pre-training stage, both the encoder and decoder are trained to reconstruct the input time series data with MSE loss. Next, in the main training stage, a conditional score matching network, based on UNet \cite{ronneberger2015u} with linear layers, is trained with the pre-trained encoder and decoder to synthesise the time series data. PC sampler \cite{song2020score} is harnessed to solve the time-reverse SDE and to yield the synthesised time series data. Its remarkable performance is demonstrated by surpassing prior baselines built upon the frameworks of VAE and GAN on five real-world datasets.

\subsubsection{Imputation}
\label{sec:timeseriesimputation}

% An explanation on CSDI
Although CSDI \cite{tashiro2021csdi} is mainly introduced for time-series data imputation, it is also applicable to forecasting. Tashiro \textit{et al}. propose novel training and sampling strategies, which are inspired by masked language modelling \cite{vaswani2017attention}. During the training process, a subset of observed values is employed as conditional data, with the rest of the observed data for imputation targets. On the other hand, during sampling, all observed data points are utilised as conditional information, whereas all missing values are considered as targets for imputation. Let conditional observations $ \mathbf{x}_0^\mathrm{co} $ and imputation targets $  \mathbf{x}_0^\mathrm{ta} $, then the noisy target to be sampled can be written as $\mathbf{x}_{t}^\mathrm{ta} = \sqrt{{\alpha}_t}\mathbf{x}_0^\mathrm{ta} +  (1-{\alpha}_t) \epsilon$. Accordingly, the imputation network $ \epsilon_{\theta} $ is trained by minimising the following objective function:

\begin{equation} \label{eq:csdi}
  \mathbb{E}_{\mathbf{x}_0 \sim p(\mathbf{x}_0), \epsilon \sim \mathcal{N}(\mathbf{0}, \mathbf{I}),t} \left [
  || (\epsilon - \epsilon_\theta (\mathbf{x}_{t}^\mathrm{ta},t \mid \mathbf{x}_{0}^\mathrm{co})) ||_2^2 \right ].
\end{equation}

To effectively harness both temporal and feature-based dependencies of time series, they design a temporal and feature Transformer layer \cite{vaswani2017attention}, instead of using a convolution layer. It outperforms other strong VAE-based baselines on both healthcare and environmental data.

% An explanation on SSSD
SSSD$^\mathrm{S4}$ \cite{alcaraz2022diffusion} is a pioneering approach that integrates structured state-space models (SSM) \cite{gu2021efficiently} into the framework of diffusion models. Specifically, SSM is proven to be effective in handling long-term dependencies on time-series data. It is formulated as:

\begin{equation} \label{eq:sssd}
	x'(t) = Ax(t) + Bu(t) \hspace*{1mm} \textrm{and} \hspace*{1mm} y(t) = Cx(t) + Du(t)
\end{equation}

where $ u(t) $ and $ y(t) $ are 1D input and output sequence, respectively, and $ x(t) $ is an $N$-dimensional hidden state. 
When learning with diffusion model framework, it denoises the segments, either in part or as a whole, making it suitable for both imputation and forecasting applications. 
First, for the imputation task, a given time series $ \mathbf{x}^0 $ and a binary mask $ \textbf{M} $ indicate observed elements, then the concatenated matrix is denoted as $ \mathbf{x}^0_{c} = concat(\mathbf{x}^0 \odot \textbf{M}, \textbf{M}) $. They propose the modification of the objective function that incorporates a binary mask, built upon the DDPM framework \cite{ho2020denoising}:

\begin{equation} \label{eq:sssd_imputation}
        \mathbb{E}_{k, \mathbf x_0, \epsilon}{ \left [ \lambda(k)  \left\| \epsilon \odot \textbf{M} - \epsilon_\theta(\sqrt{\bar{\alpha_k}} \mathbf{x}_0 + \sqrt{1 - \bar{\alpha_k}} \epsilon, \mathbf{x}^0_{c}, k) \odot \textbf{M} \right\|^2 \right] } 
\end{equation}

where $ \epsilon_\theta $ is the SSM and $k$ is the diffusion step. In forecasting, $ \mathbf{x}^{0} $ is replaced with $ (1 - \textbf{M}) \odot \mathbf{x}^{0} $. SSSD$^\mathrm{S4}$ employs SSSD and modifies the previous works \cite{tashiro2021csdi, kong2020diffwave} to form a more direct representation of a diffusion process along the time axis. It has shown its competitiveness compared to most of the prior works on handling long sequence time series.

\subsubsection{Forecasting}
\label{sec:timeseriesforecasting}

% An explanation on TimeGrad
TimeGrad, presented by Rasul \textit{et al}. \cite{rasul2021autoregressive}, is an autoregressive multivariate probabilistic time series forecasting model on the basis of DDPM framework \cite{ho2020denoising}. The time interval for prediction is perturbed using Gaussian noise, which is then gradually removed, conditioned on historical time series data. The historical data is encoded with a conditional distribution approximation using a hidden state, generated by an RNN-based module. At time point $ t $, given the multivariate time series vector $ \mathbf{x}_{t}^0 $, the recurrent model $ \mathrm{RNN} $ encodes the time series using covariates $ \mathbf{c}_{t} $ and the hidden state $ \mathbf{h}_{t-1} $ from the previous step $ t-1 $ by:
 
\begin{equation} \label{eq:timegrad_1}
    \mathbf{h}_t = \mathrm{RNN}_\theta(\mathbf{x}_{t}^0, \mathbf{c}_{t}, \mathbf{h}_{t-1}).
\end{equation}

Then, the conditional distribution is approximated as:

\begin{equation} \label{eq:timegrad_2}
    \prod_{t=t_0}^T p_\theta(\mathbf{x}_{t}^0 | \mathbf{h}_{t-1}).
\end{equation}

They follow a similar derivation introduced in Section \ref{sec:backgrounds_ddpms}, yielding an appropriate objective function using equation \ref{eq:timegrad_2}. After training, the iterative sampling process is conducted until the desired forecast length is attained.

% An explanation on ScoreGrad
ScoreGrad \cite{yan2021scoregrad} is built upon SDE framework \cite{song2020score} for time series foreceasting. It comprises a feature extraction module and conditional score-matching module. The feature extraction module, as well as target distribution, is identical to the previous work of TimeGrad \cite{rasul2021autoregressive} in that it encodes historical data and generates hidden states. Next, the conditional score-matching module has multiple residual blocks, each comprising a bidirectional dilated convolution layer, a gated activation unit, and a 1D convolutional neural network for generating output. PC sampler \cite{song2020score} is employed to forecast the future time datapoints by solving the time-reverse SDE.

% An explanation on D3VAE
Li \textit{et al}. \cite{li2022generative} propose D$^{3}$VAE, where a series of diffusion, denoising, and disentanglement is applied to bidirectional VAE (BVAE) \cite{vahdat2020nvae} with the aim of improving both performance and interpretability in time series forecasting. First, to improve performance, it harnesses a coupled forward diffusion process for data augmentation on both input and target data. It reduces both epistemic and aleatoric uncertainties, where the former is induced by the model and the latter by the data. Meanwhile, the backward process, which encompasses prediction and further refinement of the disturbed prediction, is facilitated by leveraging the BVAE and multi-scale denoising score matching. It further effectively cleans the generated time series using a single-step gradient denoising jump \cite{saremi2019neural}. 

Next, interpretability involves discerning the independent factors within the data \cite{kim2018disentangling}. It can be accomplished by disentangling the latent variables, which signify trends or seasonality. In this regard, D$^{3}$VAE \cite{li2022generative} minimises the Total Correlation (TC) \cite{kim2018disentangling} of the BVAE to disentangle latent variables. Therefore, its optimisation process is guided by a combination of four loss terms: 1) KL divergence between the estimated target distribution and target distribution, 2) denoising score matching, 3) minimisation of TC, and 4) MSE loss between prediction and ground truth. It surpasses previous VAE-based works by a significant margin.

% An explanation on DSPD/CSPD
Bilov\v s \textit{et al}. \cite{bilovs2022modeling} approach time series diffusion modelling differently from the prior works, assuming that the time series data can be represented as a series of values derived from the underlying continuous function $ \mathbf{x}( \cdot ) $. They propose a novel score-based generative diffusion framework where noise injection and removal processes are performed on the entire continuous function, instead of being applied to individual data points. In other words, the continuous function transitions to the prior stochastic process during the forward process, whilst the reverse process yields the new function samples. This means that this continuous function should be continuous and computationally tractable so as to facilitate both training and sampling. Bilov\v s \textit{et al}. \cite{bilovs2022modeling} fulfill these requirements by designing a Gaussian stochastic process with a covariate $ \Sigma $ as $ \epsilon ( \cdot ) \sim \mathcal{GP}(\mathbf{0}, \Sigma) $, instead of the conventional noise vector $ \epsilon \sim \mathcal{N} (\mathbf{0}, \mathbf{I}) $.

They first introduce Discrete Stochastic Process Diffusion (DSPD), which is built upon the DDPM formulation \cite{ho2020denoising}. The transition kernels in the forward process and backward process are modified from the original formulations of equations \ref{eq:ddpm_forward3} and \ref{eq:ddpm_backward2} to equations \ref{eq:dspd_forward} and \ref{eq:dspd_reverse}, respectively, as following:

\begin{equation} \label{eq:dspd_forward}
    q(\mathbf{X}^{k}|\mathbf{X}^0) = \mathcal{N}(\sqrt{\bar{\alpha}_{k}} \mathbf{X}^0, (1 - \bar{\alpha}_{k}) \mathbf{\Sigma}),
\end{equation}

\begin{equation} \label{eq:dspd_reverse}
    p_{\mathbf{\theta}} (\mathbf{X}^{k-1}|\mathbf{X}^{k}) = \mathcal{N}(\mu_{\mathbf{\theta}} (\mathbf{X}^{k}, {k}),(1-\alpha_{k}) \mathbf{\Sigma})
\end{equation}

where $ \mathbf{X}^{k} = (\mathbf{x}^{k}(t_0), ... , \mathbf{x}^{k}(t_{M-1})) $  represents the time-indexed sequence of points observed at $M$ distinct timestamps, $t_i$, within the interval $t \subset [0, T]$, at $ k $-th diffusion step. As aforementioned, these points are assumed to be derived from the continuous function $\mathbf{x}( \cdot )$. Next, the objective function is modified from equation \ref{eq:ddpm_ho} to \ref{eq:dspd_objective},

\begin{equation} \label{eq:dspd_objective}
    \mathbb{E}_{k, \mathbf{X}^0_{F}, \epsilon} \left[ \lambda(k) \left\| \epsilon - \epsilon_{\theta} \left(\sqrt{\bar{\alpha}_k} \mathbf{X}^0_{F} + \sqrt{1-\bar{\alpha}_k} \mathbf{\epsilon}_{k}, \mathbf{X}^0_{H}, k \right)\right\|^2 \right]
\end{equation}

where $ \mathbf{X}^0_{F} $ and $ \mathbf{X}^0_{H} $ denote the input future and historical data. The DSPD can be applied to both forecasting and imputation tasks. First, the forecasting methodology of DSPD is similar to TimeGrad \cite{rasul2021autoregressive}, with respect to objective function, architecture, and the sampling process. However, it advances in two points: 1) it is capable of predicting any future point within a continuous time interval and 2) it facilitates simultaneous prediction of multiple time points in a single run, not in an autoregressive fashion. Second, DSPD assumes that observed time series are formed by the values of the continuous function $\mathbf{x}( \cdot )$ at specific time points, which allows missing values to be computed by extracting the value of the continuous function at the corresponding time point. Thus, it can be repurposed for imputation by replacing the variance term with a covariance matrix, which is indicated in the equation on neural network of CSDI \cite{tashiro2021csdi}. 

Bilov\v s \textit{et al}. \cite{bilovs2022modeling} extend DSPD to a continuous variant termed as Continuous Stochastic Process Diffusion (CSPD), developed from the SDE formulation \cite{song2020score}. This continuous variant can also be employed for both imputation and forecasting tasks, particularly useful in imputation tasks. This is because the continuous noise process is more natural when handling irregular time intervals, which is a prevalent issue in imputation tasks.

\section{Domain-specific Applications}
\label{sec:domain}

\subsection{Electronic Health Records (EHR)}
\label{sec:ehr}

% An overview on EHR
EHR encompasses a vast amount of patient-centered information, including patient's medical history, diagnoses, medications and such. It has significantly benefited from the advancements in deep learning applications, such as predictive diagnosis \cite{li2020behrt}, medication recommendation \cite{bhoi2021personalizing}, and continuous monitoring in intensive care units (ICUs), which significantly diminished the likelihood of complications and mortality rates \cite{poncette2019clinical}. However, due to the privacy and ethical concerns surrounding the EHR, release of public data has been limited, which has constrained further research and development \cite{naik2022legal}. To mitigate these concerns, high-quality and realistic EHR synthesis using GAN \cite{baowaly2019synthesizing} has been investigated amongst researchers. With more powerful generative diffusion models, researchers have explored both synthesis and forecasting of the EHR.

\subsubsection{Generation}
\label{sec:ehrgeneration}
 
% An explanation on MedDiff
MedDiff \cite{he2023meddiff} is the first literature that applies diffusion models to EHR. He \textit{et al}. \cite{he2023meddiff} propose novel three points: first, it accelerates the generation process via Anderson acceleration \cite{anderson1965iterative}. Next, to reflect label information to the generated samples, it utilises classifier-guided sampling process \cite{dhariwal2021diffusion}. Lastly, it uses 1D convolutional layer to U-Net architecture to enhance the learning of feature correlations amongst neighbor features. Nonetheless, it is still limited in that it only generates continuous variables, as the authors confine the data to follow Gaussian distribution.

% An explanation on TabDDPM
To mitigate the aforementioned issue, Ceritli \textit{et al}. \cite{ceritli2023synthesizing} adopt TabDDPM framework \cite{kotelnikov2022tabddpm} to generate 1) continuous data using the Gaussian multinomial diffusion processes and 2) categorical data using the multinomial diffusion processes. The results demonstrate that their work \cite{ceritli2023synthesizing} surpasses previous works in data usability criteria but with the exception of privacy evaluation, raising the necessity to improve in such direction.

% An explanation on EHRDiff
EHRDiff \cite{yuan2023ehrdiff} is built upon the work of SDE \cite{song2020score}, especially solving the reverse process via Heun's second order method ODEs to produce more precise and realistic EHR samples in a deterministic manner. Also, it utilises an adaptive parameterisation \cite{karras2022elucidating} to address the issue of amplified prediction error due to the variance of scale noise $ \sigma $. For the architecture, it uses multilayer perceptron (MLP) layers. Similar to TabDDPM \cite{ceritli2023synthesizing}, it shows outstanding performance on data usability, but moderate performance on privacy criteria.

% An explanation on EHR-DPM
Kuo \textit{et al}. \cite{kuo2023synthetic} propose a EHR-DPM. It introduces two auxiliary loss functions: 1) one-step reconstruction loss to alleviate training instability and 2) MSE loss between the clean data and its restored counterpart in a randomly projected latent space. Unlike MedDiff \cite{he2023meddiff} that applies a 1D convolutional layer to individually denoise each feature, EHR-DPM adds linear layers. It involves two steps: 1) a linear layer is added to perform denoising at the latent level, as opposed to the variable level, 2) an additional linear layer is incorporated into each up-sampling and down-sampling 1D CNN, and the final up-sampling output. 

\subsubsection{Forecasting}
\label{sec:ehrforecasting}

% An explanation on TDSTF
TDSTF \cite{chang2023tdstf} is an initial work on EHR forecasting via diffusion models. The training and sampling process are the same as other traditional diffusion models for time series forecasting. However, to overcome data sparsity, Chang \textit{et al}. \cite{chang2023tdstf} convert the data into triplet form instead of utilising conventional techniques such as aggregation and imputation. Also, its architecture consists of multiple residual layers including Transformer encoder and decoder \cite{vaswani2017attention}. It surpasses previous existing models in terms of both predictive performance on critical vital sign and inference speed by a large margin on MIMIC-III dataset \cite{johnson2016mimic}.

\subsection{Bioelectrical Signal Processing}
\label{sec:bioelectricalsignals}

% An overview on bioelectrical signals
Bioelectrical signals are measured through electrical potential differences across a cell or an organ \cite{pal2015evaluation}. They include Electrocardiogram (ECG), Electroencephalogram (EEG), Electromyogram (EMG) and Electrooculogram (EOG). These signals, expressing electrical activity in the heart, brain, muscles, and eyes, respectively, serve as non-invasive but informative diagnostic tools \cite{zimetbaum2003use}. 
Recent advancements in deep learning methodologies have led to notable predictive capability improvements on bioelectrical signals. However, two significant hurdles persist \cite{chatterjee2020review}: restricted availability of public bioelectrical signal data due to privacy concerns and inevitable data corruption owing to noise from sources like patient respiration and body movements. These challenges impede further improvements of deep predictive models, thus researchers have explored deep generative models to both synthesise \cite{delaney2019synthesis} and reduce wander or noise \cite{romero2021deepfilter} from bioelectrical signal data. 

\subsubsection{Generation} 
\label{sec:bioelectricalsignalsgeneration}

% An explanation on SSSD-ECG
SSSD-ECG \cite{alcaraz2023diffusion} is built upon the work of SSSD$^\mathrm{S4}$ \cite{alcaraz2022diffusion}, which integrates diffusion models and structured state space models. It synthesises 12-lead ECG data in a multi-label fashion by conditioning on over 70 ECG statements. It outperforms previous GAN-based works in ECG synthesis in terms of both quantitative evaluation and qualitative analysis from domain experts. 

\subsubsection{Enhancement} 
\label{sec:bioelectricalsignalsenhancement}

% An explanation on DeScoD-ECG
Li \textit{et al}. \cite{li2023descod} propose a novel ECG enhancement model, DeScoD-ECG, that is designed for the removal of noise and wander. It operates conditionally on noisy observations and employs an iterative process to restore signals from Gaussian distributions. With a clean input signal $ \tilde{\mathbf{x}} $ and a noisy ECG signal $ \mathbf{x}_0 $, the equation \ref{eq:ddpm_backward1} and \ref{eq:ddpm_backward2} from DDPM \cite{ho2020denoising} is modified as:

\begin{equation} \label{eq:descod_ecg_1}
    p_{\theta}\left(\mathbf{x}_{0: T}\mid \tilde{\mathbf{x}}\right)=p\left(\mathbf{x}_{T}\right) \prod_{t=1}^{T} p_{\theta}\left(\mathbf{x}_{t-1} \mid \mathbf{x}_{t},\tilde{\mathbf{x}}\right) \hspace*{1mm} \textrm{where} \hspace*{1mm}  \mathbf{x}_{T} \sim \mathcal{N}(\mathbf{0}, \mathbf{I}),
\end{equation}

\begin{equation} \label{eq:descod_ecg_2}
    p_{\theta}\left(\mathbf{x}_{t-1} \mid \mathbf{x}_{t},\tilde{\mathbf{x}}\right)=\mathcal{N}\left(\mathbf{x}_{t-1} ; \boldsymbol{\mu}_{\theta}\left(\mathbf{x}_{t}, t\mid\tilde{\mathbf{x}}\right), \sigma_{\theta}\left(\mathbf{x}_{t}, t\mid\tilde{\mathbf{x}}\right) \mathbf{I}\right).
\end{equation}

Also, it utilises a self-ensemble strategy, wherein it averages multiple output results to enhance the performance of signal reconstruction. The architecture has two backbones that extract features from both the noisy observations and the latent variables, then they are combined through bridge modules. Inspired by the DeepFilter \cite{romero2021deepfilter}, two backbones conduct multi-scale feature aggregation and channel-wise concatenation processes. Also, the bridge modules are conditioned on noise level and perform encoding using sinusoidal positional embeddings \cite{vaswani2017attention}, followed by a convolutional layer with a 1 x 1 kernel size. The absence of any up/downsampling procedures within the architecture ensures its applicability to ECG signals of any length. DeScoD-ECG \cite{li2023descod} is validated on real-world datasets, including the QT database \cite{laguna1997database} and the MIT-BIH Noise Stress Test Database \cite{moody1984noise}. The results demonstrate the efficacy of stable DeScoD-ECG compared to other baselines, particularly in situations with extreme corruption. 

% An explanation on DS-DDPM
Duan \textit{et al}. \cite{duan2023domain} explore the reconstruction of EEG signal under subject-specific variability. They propose DS-DDPM based on the assumption that noisy EEG signals (input) can be decomposed into domain specific noise and clean signals. Thus, they propose a novel diffusion and the reversal process, in which the domain specific noise and clean signal are divided and aggregated at every time step. As both noise and signal exist in orthogonal spaces, DS-DDPM learns using an Additive Angular Margin classification (Arc-Margin) loss \cite{deng2019arcface} to improve intra-class cohesion and inter-class discrepancy. Additionally, they implement an input overlap segmentation strategy to minimise temporal differences in overlapping segments. It also adopts a classifier guidance strategy \cite{dhariwal2021diffusion} that exploits human subject index to improve generalisation capability. It learns with modified UNet architecture \cite{ronneberger2015u} with multi-head attention \cite{vaswani2017attention}. The effectiveness of DS-DDPM is validated on the BCI Competition IV motor imagery decoding dataset \cite{tangermann2012review}. It is also tested on downstream classification tasks to corroborate its reconstruction capability.

\subsection{Recommendation Systems (RecSys)}
\label{sec:recsys}

% An overview on recommendation systems
RecSys aim at modelling personalised users' preferences based on their previous user-item interactions such as clicks, ratings, or purchases \cite{zhang2021artificial}. Due to their profound significant value in industry, research on RecSys has progressed from traditional methods such as collaborative filtering (CF) based techniques \cite{he2016fast} to deep learning methodologies \cite{sun2019bert4rec}. However, they often show limited generalisation performance, on account of weak collaborative signals, inadequate latent representations, or noisy data scenarios \cite{du2023sequential}. Researchers have explored the generative models of VAE \cite{liang2018variational} and GAN \cite{ren2020sequential} to mitigate these challenges, but these models also have their own limitations of restricted capability in capturing personalised user preferences (from VAE-based models) and training instability (from GAN-based models) \cite{wang2023diffusion}. To address the issues, researchers have delved into diffusion models, owing to their strong representation capability and training stability. 

% An explanation on CODIGEM
Walker \textit{et al}. \cite{walker2022recommendation} propose CODIGEM, which is the first work exploiting diffusion models on RecSys. It generates strong collaborative signals and robust latent representations on user-item interactions, thus outperforming previous VAE-based works. However, the model falls short in handling sequential scenarios.

% An explanation on DiffuRec
DiffuRec \cite{li2023diffurec} aims to address the limitations from modelling the item representations as fixed vectors. To handle latent representation of items and multi-level interests of users, they utilise diffusion models to represent them as distributions. During the diffusion process, a truncated linear schedule is used for noise addition. The introduction of noise functions as an uncertainty factor that steers the learning process towards enhanced robustness. Also, instead of using conventional MSE loss in DDPM \cite{ho2020denoising}, it learns with cross-entropy loss. This modification is driven by two reasons: the static nature of item embedding and the use of inner product for relevance calculation in the reverse process. They also exploit Transformer \cite{vaswani2017attention} as an approximator to reconstruct target item representation. These modifications enable their model to show remarkable performance on several benchmark datasets.

% An explanation on DiffRec (Du et al.)
Du \textit{et al}. \cite{du2023sequential} develop another novel sequential diffusion model for RecSys. It introduces an additional transition to convert the items with discrete features. It perturbs the original item, instead of the whole item sequences, by injecting noise and it restores the target by MSE loss term. Also, it samples only important diffusion steps instead of the entire steps. Those strategies enable an efficient training and sampling. On the architectural side, it leverages a transformer-based encoder \cite{vaswani2017attention} with learnable positional embedding. It significantly surpasses preceding generative and contrastive learning based benchmarks on real world datasets.

% An explanation on DiffRec (Wang et al.)
Wang \textit{et al}. \cite{wang2023diffusion} design two novel diffusion recommendation models for sequential RecSys of L-DiffRec and T-DiffRec, highlighting two major challenges in RecSys: 1) high computational costs for large-scale item prediction and 2) temporal transition for user preference. L-DiffRec addresses the first challenge. It groups items into clusters, then compresses the user-item interaction using multiple VAEs and conducts diffusion processes in the latent space to generate top-K recommendations. It addresses the high computational demands arising from large-scale item prediction. 

Second, T-DiffRec mitigates the second issue, temporal shifts in user preferences. It introduces time-aware reweighting strategy to model user interactions based on the assumption that the recent interactions of a user may capture user preference more effectively. This enhances the model's adaptability to dynamic user behaviour. The two variants of DiffRec outperform previous VAE-based baselines by a substantial margin on several public datasets.

% An explanation on CDDRec
CDDRec \cite{wang2023conditional} achieves diffusion models for sequential RecSys in conditional autoregressive and ranking aware manners. Its network comprises of three models: a step-wise diffuser, a sequence encoder and a cross-attentive conditional denoising decoder. First, the step-wise diffuser corrupts the original data using increasing Gaussian noise. The sequence encoder utilises a self-attention mechanism on historical interactions whilst the cross-attentive conditional denoising decoder adopts a direct-condition mechanism wherein the sequence embeddings from the encoder are directly conditioned to the decoder at every denoising step. It enables a conditional sample generation in an autoregressive fashion. 

Further, to mitigate the inherent long-tailed and sparse item distribution, CDDRec \cite{wang2023conditional} is optimised through a combination of a cross-divergence loss and multi-view contrastive loss with denoising diffusion loss. The cross-divergence loss term utilises KL divergence to minimise the discrepancy between the forecasted mean and the target item embedding, as compared to the discrepancy between the predicted mean and unrelated target embeddings. It helps avoid a situation where the rank scores uniformly converge across all items for all users, enabling the model to function in a ranking-aware fashion. The multi-view contrastive loss with denoising diffusion loss applies contrastive loss to enhance the robustness against the noisy interference by maximising the agreement between the original view and its counterpart, achieved through random cropping, shuffling and masking randomly. The combination of two loss terms, therefore, guides the CDDRec to be more discriminative and robust.

\section{Current Challenges and Future Research Directions}
\label{sec:challengesnopportunities}

% Overview
Despite the astonishing advancements in generative diffusion models for the structured data, there are still challenges that this field continues to grapple with. The objective of this section is to describe these challenges and provide potential future research directions.

% Customised Designs for Structured Data
\textbf{Customised Designs for Structured Data} Generative diffusion models are largely influenced by factors such as architectural design, training strategy, and noise scheduling. However, most of the approaches on structured data have directly adopted or slightly modified from existing seminal works from other data modalities. Customising these elements may potentially lead to significant improvements in modelling capability and overcoming generative modelling trilemma: sampling quality, diversity, and speed \cite{xiao2022tackling}. 

% Causal Learning and Counterfactual Reasoning
\textbf{Causal Learning and Counterfactual Reasoning} Causal learning aims to discern and identify the casual relationships or inter-dependencies amongst variables within a given dataset \cite{moraffah2021causal}. For instance, it allows us to model potential outcomes or the most indicative factor in financial forecasting or disease progression. On the other hand, counterfactual reasoning aims to predict an individual's outcome under different circumstances \cite{liu2022practical}. Specifically, it predicts what the result would be like if certain variables were changed from their observed real-world values. Hence, the integration of causal learning or counterfactual reasoning methodologies into the diffusion models can potentially enhance their performance. By harnessing cause-and-effect relationships or counterfactual estimation rather than simply exploiting correlations or conditioning on variables, we can optimise diffusion models to be more reliable and robust.

% Bias in Dataset
\textbf{Bias in Dataset} Another noteworthy challenge is the inherent bias present within the publicly available datasets. Specifically, the demographic features in the EHRs and the bioelectrical signals extracted from the subjects are often biased towards specific classes \cite{duan2023domain, meng2022interpretability}. This inevitable skewness in data consequently impacts the generalisability of the generative models and limits their applicability to related fields. To circumvent these issues, it is vital to utilise more diverse and balanced datasets. Otherwise, novel methods should be developed to neutralise this inherent bias.

% Extensions to Multi-modality Learning
\textbf{Extensions to Multi-modality Learning} The fusion of structured data with other data modalities not only enhances the model performance but also expands the possible tasks, \textit{e.g.,} improving financial stock price prediction by jointly learning both text and time series data \cite{muthukumar2021stochastic} and text synthesis from table \cite{liu2018table}. In this regard, future research direction may focus on the development of novel methodologies adept at integrating multiple data modalities in an efficient and effective manner as well as exploring the potential untapped tasks.

% Miscellany 
\textbf{Miscellany} In the realm of tabular data modelling, it is possible to devise an enhanced strategy that effectively models both continuous and categorical data types. Furthermore, a universal framework can be developed to encompass generation, imputation, and forecasting for both tabular and time-series modelling. For applications involving EHR and biological signal processing, current works may benefit from an incorporation of domain-specific knowledge or issues, \textit{e.g.,} medical ontology and irregular visit interval \cite{niu2022fusion}. Furthermore, the long inference time compared to GAN-based methods poses a challenge for real-time or on-chip deployment in medical equipment, highlighting the need for improvements to facilitate practical use.

\section{Conclusion} 
\label{sec:conclusion}

% Conclusion
The generative diffusion models have verified themselves by exhibiting remarkable performance across diverse applications, and also have shown excellent performance on structured data. However, current research on generative diffusion models specifically tailored for structured data has not received active attention, compared to the other data modalities. To facilitate exploration and advancement in this field, we provide a comprehensive survey on generative diffusion models for structured data. Our survey encompasses a brief introduction to the underlying theory of score-based diffusion models, followed by a concise review of existing literature categorized into data-driven general tasks and domain-specific applications. Additionally, we discuss current challenges and outline potential research directions for the future. We hope that this survey will serve as a valuable guide for those interested in this field, thereby fostering further research and advancement in the area.

\bibliographystyle{unsrt}
\bibliography{main}

\begin{thebibliography}{100}

\bibitem{lin2021forecasting}
Yu~Lin, Yan Yan, Jiali Xu, Ying Liao, and Feng Ma.
\newblock Forecasting stock index price using the ceemdan-lstm model.
\newblock {\em The North American Journal of Economics and Finance}, 57:101421,
  2021.

\bibitem{tiwari2021credit}
Pooja Tiwari, Simran Mehta, Nishtha Sakhuja, Jitendra Kumar, and Ashutosh~Kumar
  Singh.
\newblock Credit card fraud detection using machine learning: a study.
\newblock {\em arXiv preprint arXiv:2108.10005}, 2021.

\bibitem{zhou2019deep}
Guorui Zhou, Na~Mou, Ying Fan, Qi~Pi, Weijie Bian, Chang Zhou, Xiaoqiang Zhu,
  and Kun Gai.
\newblock Deep interest evolution network for click-through rate prediction.
\newblock In {\em Proceedings of the AAAI conference on artificial
  intelligence}, volume~33, pages 5941--5948, 2019.

\bibitem{beunza2019comparison}
Juan-Jose Beunza, Enrique Puertas, Ester Garc{\'\i}a-Ovejero, Gema Villalba,
  Emilia Condes, Gergana Koleva, Cristian Hurtado, and Manuel~F Landecho.
\newblock Comparison of machine learning algorithms for clinical event
  prediction (risk of coronary heart disease).
\newblock {\em Journal of biomedical informatics}, 97:103257, 2019.

\bibitem{liu2022practical}
Licheng Liu, Ye~Wang, and Yiqing Xu.
\newblock A practical guide to counterfactual estimators for causal inference
  with time-series cross-sectional data.
\newblock {\em American Journal of Political Science}, 2022.

\bibitem{Kim2021AGCNAG}
Seunghoi Kim and Daniel~C. Alexander.
\newblock Agcn: Adversarial graph convolutional network for 3d point cloud
  segmentation.
\newblock In {\em British Machine Vision Conference}, 2021.

\bibitem{song2021solving}
Yang Song, Liyue Shen, Lei Xing, and Stefano Ermon.
\newblock Solving inverse problems in medical imaging with score-based
  generative models.
\newblock In {\em NeurIPS 2021 Workshop on Deep Learning and Inverse Problems},
  2021.

\bibitem{anuchitanukul2022surf}
Atijit Anuchitanukul and Julia Ive.
\newblock Surf: Semantic-level unsupervised reward function for machine
  translation.
\newblock In {\em Proceedings of the 2022 Conference of the North American
  Chapter of the Association for Computational Linguistics: Human Language
  Technologies}, pages 4508--4522, 2022.

\bibitem{kim2022multi}
To~Eun Kim and Aldo Lipani.
\newblock A multi-task based neural model to simulate users in goal oriented
  dialogue systems.
\newblock In {\em Proceedings of the 45th International ACM SIGIR Conference on
  Research and Development in Information Retrieval}, pages 2115--2119, 2022.

\bibitem{Kim_2022_CVPR}
Gwanghyun Kim, Taesung Kwon, and Jong~Chul Ye.
\newblock Diffusionclip: Text-guided diffusion models for robust image
  manipulation.
\newblock In {\em Proceedings of the IEEE/CVF Conference on Computer Vision and
  Pattern Recognition (CVPR)}, pages 2426--2435, June 2022.

\bibitem{gu2022vector}
Shuyang Gu, Dong Chen, Jianmin Bao, Fang Wen, Bo~Zhang, Dongdong Chen, Lu~Yuan,
  and Baining Guo.
\newblock Vector quantized diffusion model for text-to-image synthesis.
\newblock In {\em Proceedings of the IEEE/CVF Conference on Computer Vision and
  Pattern Recognition}, pages 10696--10706, 2022.

\bibitem{li2022diffusion}
Xiang Li, John Thickstun, Ishaan Gulrajani, Percy~S Liang, and Tatsunori~B
  Hashimoto.
\newblock Diffusion-lm improves controllable text generation.
\newblock {\em Advances in Neural Information Processing Systems},
  35:4328--4343, 2022.

\bibitem{condita2022alexaprize}
Jerome Ramos, To~Eun Kim, Zhengxiang Shi, Xiao Fu, Fanghua Ye, Yue Feng, and
  Aldo Lipani.
\newblock Condita: A state machine like architecture for multi-modal task bots.
\newblock In {\em Alexa Prize TaskBot Challenge Proceedings}, 2022.

\bibitem{yu2022latent}
Peiyu Yu, Sirui Xie, Xiaojian Ma, Baoxiong Jia, Bo~Pang, Ruigi Gao, Yixin Zhu,
  Song-Chun Zhu, and Ying~Nian Wu.
\newblock Latent diffusion energy-based model for interpretable text modeling.
\newblock {\em arXiv preprint arXiv:2206.05895}, 2022.

\bibitem{chung2023diffusion}
Hyungjin Chung, Jeongsol Kim, Michael~Thompson Mccann, Marc~Louis Klasky, and
  Jong~Chul Ye.
\newblock Diffusion posterior sampling for general noisy inverse problems.
\newblock In {\em The Eleventh International Conference on Learning
  Representations}, 2023.

\bibitem{lecun2015deep}
Yann LeCun, Yoshua Bengio, and Geoffrey Hinton.
\newblock Deep learning.
\newblock {\em nature}, 521(7553):436--444, 2015.

\bibitem{elsayed2021we}
Shereen Elsayed, Daniela Thyssens, Ahmed Rashed, Hadi~Samer Jomaa, and Lars
  Schmidt-Thieme.
\newblock Do we really need deep learning models for time series forecasting?
\newblock {\em arXiv preprint arXiv:2101.02118}, 2021.

\bibitem{shwartz2022tabular}
Ravid Shwartz-Ziv and Amitai Armon.
\newblock Tabular data: Deep learning is not all you need.
\newblock {\em Information Fusion}, 81:84--90, 2022.

\bibitem{kadra2021well}
Arlind Kadra, Marius Lindauer, Frank Hutter, and Josif Grabocka.
\newblock Well-tuned simple nets excel on tabular datasets.
\newblock {\em Advances in neural information processing systems},
  34:23928--23941, 2021.

\bibitem{kingma2013auto}
Diederik~P Kingma and Max Welling.
\newblock Auto-encoding variational bayes.
\newblock {\em arXiv preprint arXiv:1312.6114}, 2013.

\bibitem{goodfellow2020generative}
Ian Goodfellow, Jean Pouget-Abadie, Mehdi Mirza, Bing Xu, David Warde-Farley,
  Sherjil Ozair, Aaron Courville, and Yoshua Bengio.
\newblock Generative adversarial networks.
\newblock {\em Communications of the ACM}, 63(11):139--144, 2020.

\bibitem{vardhan2020generating}
L~Vivek~Harsha Vardhan and Stanley Kok.
\newblock Generating privacy-preserving synthetic tabular data using oblivious
  variational autoencoders.
\newblock In {\em Proceedings of the Workshop on Economics of Privacy and Data
  Labor at the 37 th International Conference on Machine Learning}, 2020.

\bibitem{esteban2017real}
Crist{\'o}bal Esteban, Stephanie~L Hyland, and Gunnar R{\"a}tsch.
\newblock Real-valued (medical) time series generation with recurrent
  conditional gans.
\newblock {\em arXiv preprint arXiv:1706.02633}, 2017.

\bibitem{wiese2020quant}
Magnus Wiese, Robert Knobloch, Ralf Korn, and Peter Kretschmer.
\newblock Quant gans: deep generation of financial time series.
\newblock {\em Quantitative Finance}, 20(9):1419--1440, 2020.

\bibitem{yoon2018gain}
Jinsung Yoon, James Jordon, and Mihaela Schaar.
\newblock Gain: Missing data imputation using generative adversarial nets.
\newblock In {\em International conference on machine learning}, pages
  5689--5698. PMLR, 2018.

\bibitem{song2019generative}
Yang Song and Stefano Ermon.
\newblock Generative modeling by estimating gradients of the data distribution.
\newblock {\em Advances in neural information processing systems}, 32, 2019.

\bibitem{ho2020denoising}
Jonathan Ho, Ajay Jain, and Pieter Abbeel.
\newblock Denoising diffusion probabilistic models.
\newblock {\em Advances in Neural Information Processing Systems},
  33:6840--6851, 2020.

\bibitem{song2020score}
Yang Song, Jascha Sohl-Dickstein, Diederik~P Kingma, Abhishek Kumar, Stefano
  Ermon, and Ben Poole.
\newblock Score-based generative modeling through stochastic differential
  equations.
\newblock {\em arXiv preprint arXiv:2011.13456}, 2020.

\bibitem{sohl2015deep}
Jascha Sohl-Dickstein, Eric Weiss, Niru Maheswaranathan, and Surya Ganguli.
\newblock Deep unsupervised learning using nonequilibrium thermodynamics.
\newblock In {\em International Conference on Machine Learning}, pages
  2256--2265. PMLR, 2015.

\bibitem{huang2022mdm}
Lei Huang, Hengtong Zhang, Tingyang Xu, and Ka-Chun Wong.
\newblock Mdm: Molecular diffusion model for 3d molecule generation.
\newblock {\em arXiv preprint arXiv:2209.05710}, 2022.

\bibitem{xu2023geometric}
Minkai Xu, Alexander Powers, Ron Dror, Stefano Ermon, and Jure Leskovec.
\newblock Geometric latent diffusion models for 3d molecule generation.
\newblock {\em arXiv preprint arXiv:2305.01140}, 2023.

\bibitem{yang2022diffusion}
Ling Yang, Zhilong Zhang, Yang Song, Shenda Hong, Runsheng Xu, Yue Zhao,
  Yingxia Shao, Wentao Zhang, Bin Cui, and Ming-Hsuan Yang.
\newblock Diffusion models: A comprehensive survey of methods and applications.
\newblock {\em arXiv preprint arXiv:2209.00796}, 2022.

\bibitem{cao2022survey}
Hanqun Cao, Cheng Tan, Zhangyang Gao, Guangyong Chen, Pheng-Ann Heng, and
  Stan~Z Li.
\newblock A survey on generative diffusion model.
\newblock {\em arXiv preprint arXiv:2209.02646}, 2022.

\bibitem{croitoru2023diffusion}
Florinel-Alin Croitoru, Vlad Hondru, Radu~Tudor Ionescu, and Mubarak Shah.
\newblock Diffusion models in vision: A survey.
\newblock {\em IEEE Transactions on Pattern Analysis and Machine Intelligence},
  2023.

\bibitem{zhu2023diffusion}
Yuansong Zhu and Yu~Zhao.
\newblock Diffusion models in nlp: A survey.
\newblock {\em arXiv preprint arXiv:2303.07576}, 2023.

\bibitem{zhang2023survey}
Mengchun Zhang, Maryam Qamar, Taegoo Kang, Yuna Jung, Chenshuang Zhang, Sung-Ho
  Bae, and Chaoning Zhang.
\newblock A survey on graph diffusion models: Generative ai in science for
  molecule, protein and material.
\newblock {\em arXiv preprint arXiv:2304.01565}, 2023.

\bibitem{kazerouni2023diffusion}
Amirhossein Kazerouni, Ehsan~Khodapanah Aghdam, Moein Heidari, Reza Azad,
  Mohsen Fayyaz, Ilker Hacihaliloglu, and Dorit Merhof.
\newblock Diffusion models for medical image analysis: A comprehensive survey,
  2023.

\bibitem{lin2023diffusion}
Lequan Lin, Zhengkun Li, Ruikun Li, Xuliang Li, and Junbin Gao.
\newblock Diffusion models for time series applications: A survey.
\newblock {\em arXiv preprint arXiv:2305.00624}, 2023.

\bibitem{tashiro2021csdi}
Yusuke Tashiro, Jiaming Song, Yang Song, and Stefano Ermon.
\newblock Csdi: Conditional score-based diffusion models for probabilistic time
  series imputation.
\newblock {\em Advances in Neural Information Processing Systems},
  34:24804--24816, 2021.

\bibitem{vignac2022digress}
Clement Vignac, Igor Krawczuk, Antoine Siraudin, Bohan Wang, Volkan Cevher, and
  Pascal Frossard.
\newblock Digress: Discrete denoising diffusion for graph generation.
\newblock {\em arXiv preprint arXiv:2209.14734}, 2022.

\bibitem{hoogeboom2021argmax}
Emiel Hoogeboom, Didrik Nielsen, Priyank Jaini, Patrick Forr{\'e}, and Max
  Welling.
\newblock Argmax flows and multinomial diffusion: Learning categorical
  distributions.
\newblock {\em Advances in Neural Information Processing Systems},
  34:12454--12465, 2021.

\bibitem{kotelnikov2022tabddpm}
Akim Kotelnikov, Dmitry Baranchuk, Ivan Rubachev, and Artem Babenko.
\newblock Tabddpm: Modelling tabular data with diffusion models.
\newblock {\em arXiv preprint arXiv:2209.15421}, 2022.

\bibitem{hyvarinen05a}
Aapo Hyv{{\"a}}rinen.
\newblock Estimation of non-normalized statistical models by score matching.
\newblock {\em Journal of Machine Learning Research}, 6(24):695--709, 2005.

\bibitem{vincent2011connection}
Pascal Vincent.
\newblock A connection between score matching and denoising autoencoders.
\newblock {\em Neural computation}, 23(7):1661--1674, 2011.

\bibitem{song2020sliced}
Yang Song, Sahaj Garg, Jiaxin Shi, and Stefano Ermon.
\newblock Sliced score matching: A scalable approach to density and score
  estimation.
\newblock In {\em Uncertainty in Artificial Intelligence}, pages 574--584.
  PMLR, 2020.

\bibitem{kim2022sos}
Jayoung Kim, Chaejeong Lee, Yehjin Shin, Sewon Park, Minjung Kim, Noseong Park,
  and Jihoon Cho.
\newblock Sos: Score-based oversampling for tabular data.
\newblock In {\em Proceedings of the 28th ACM SIGKDD Conference on Knowledge
  Discovery and Data Mining}, pages 762--772, 2022.

\bibitem{kim2022stasy}
Jayoung Kim, Chaejeong Lee, and Noseong Park.
\newblock Stasy: Score-based tabular data synthesis.
\newblock {\em arXiv preprint arXiv:2210.04018}, 2022.

\bibitem{lee2023codi}
Chaejeong Lee, Jayoung Kim, and Noseong Park.
\newblock Codi: Co-evolving contrastive diffusion models for mixed-type tabular
  synthesis.
\newblock {\em arXiv preprint arXiv:2304.12654}, 2023.

\bibitem{zheng2022diffusion}
Shuhan Zheng and Nontawat Charoenphakdee.
\newblock Diffusion models for missing value imputation in tabular data.
\newblock {\em arXiv preprint arXiv:2210.17128}, 2022.

\bibitem{lim2023regular}
Haksoo Lim, Minjung Kim, Sewon Park, and Noseong Park.
\newblock Regular time-series generation using sgm.
\newblock {\em arXiv preprint arXiv:2301.08518}, 2023.

\bibitem{alcaraz2022diffusion}
Juan Miguel~Lopez Alcaraz and Nils Strodthoff.
\newblock Diffusion-based time series imputation and forecasting with
  structured state space models.
\newblock {\em arXiv preprint arXiv:2208.09399}, 2022.

\bibitem{bilovs2022modeling}
Marin Bilo{\v{s}}, Kashif Rasul, Anderson Schneider, Yuriy Nevmyvaka, and
  Stephan G{\"u}nnemann.
\newblock Modeling temporal data as continuous functions with process
  diffusion.
\newblock {\em arXiv preprint arXiv:2211.02590}, 2022.

\bibitem{rasul2021autoregressive}
Kashif Rasul, Calvin Seward, Ingmar Schuster, and Roland Vollgraf.
\newblock Autoregressive denoising diffusion models for multivariate
  probabilistic time series forecasting.
\newblock In {\em International Conference on Machine Learning}, pages
  8857--8868. PMLR, 2021.

\bibitem{yan2021scoregrad}
Tijin Yan, Hongwei Zhang, Tong Zhou, Yufeng Zhan, and Yuanqing Xia.
\newblock Scoregrad: Multivariate probabilistic time series forecasting with
  continuous energy-based generative models.
\newblock {\em arXiv preprint arXiv:2106.10121}, 2021.

\bibitem{li2022generative}
Yan Li, Xinjiang Lu, Yaqing Wang, and Dejing Dou.
\newblock Generative time series forecasting with diffusion, denoise, and
  disentanglement.
\newblock {\em Advances in Neural Information Processing Systems},
  35:23009--23022, 2022.

\bibitem{yuan2023ehrdiff}
Hongyi Yuan, Songchi Zhou, and Sheng Yu.
\newblock Ehrdiff: Exploring realistic ehr synthesis with diffusion models.
\newblock {\em arXiv preprint arXiv:2303.05656}, 2023.

\bibitem{ceritli2023synthesizing}
Taha Ceritli, Ghadeer~O Ghosheh, Vinod~Kumar Chauhan, Tingting Zhu, Andrew~P
  Creagh, and David~A Clifton.
\newblock Synthesizing mixed-type electronic health records using diffusion
  models.
\newblock {\em arXiv preprint arXiv:2302.14679}, 2023.

\bibitem{he2023meddiff}
Huan He, Shifan Zhao, Yuanzhe Xi, and Joyce~C Ho.
\newblock Meddiff: Generating electronic health records using accelerated
  denoising diffusion model.
\newblock {\em arXiv preprint arXiv:2302.04355}, 2023.

\bibitem{kuo2023synthetic}
Nicholas~I Kuo, Louisa Jorm, Sebastiano Barbieri, et~al.
\newblock Synthetic health-related longitudinal data with mixed-type variables
  generated using diffusion models.
\newblock {\em arXiv preprint arXiv:2303.12281}, 2023.

\bibitem{chang2023tdstf}
Ping Chang, Huayu Li, Stuart~F Quan, Janet Roveda, and Ao~Li.
\newblock Tdstf: Transformer-based diffusion probabilistic model for sparse
  time series forecasting.
\newblock {\em arXiv preprint arXiv:2301.06625}, 2023.

\bibitem{alcaraz2023diffusion}
Juan Miguel~Lopez Alcaraz and Nils Strodthoff.
\newblock Diffusion-based conditional ecg generation with structured state
  space models.
\newblock {\em arXiv preprint arXiv:2301.08227}, 2023.

\bibitem{li2023descod}
Huayu Li, Gregory Ditzler, Janet Roveda, and Ao~Li.
\newblock Descod-ecg: Deep score-based diffusion model for ecg baseline wander
  and noise removal.
\newblock {\em IEEE Journal of Biomedical and Health Informatics}, 2023.

\bibitem{duan2023domain}
Yiqun Duan, Jinzhao Zhou, Zhen Wang, Yu-Cheng Chang, Yu-Kai Wang, and Chin-Teng
  Lin.
\newblock Domain-specific denoising diffusion probabilistic models for brain
  dynamics.
\newblock {\em arXiv preprint arXiv:2305.04200}, 2023.

\bibitem{walker2022recommendation}
Joojo Walker, Ting Zhong, Fengli Zhang, Qiang Gao, and Fan Zhou.
\newblock Recommendation via collaborative diffusion generative model.
\newblock In {\em Knowledge Science, Engineering and Management: 15th
  International Conference, KSEM 2022, Singapore, August 6--8, 2022,
  Proceedings, Part III}, pages 593--605. Springer, 2022.

\bibitem{li2023diffurec}
Zihao Li, Aixin Sun, and Chenliang Li.
\newblock Diffurec: A diffusion model for sequential recommendation.
\newblock {\em arXiv preprint arXiv:2304.00686}, 2023.

\bibitem{du2023sequential}
Hanwen Du, Huanhuan Yuan, Zhen Huang, Pengpeng Zhao, and Xiaofang Zhou.
\newblock Sequential recommendation with diffusion models.
\newblock {\em arXiv preprint arXiv:2304.04541}, 2023.

\bibitem{wang2023diffusion}
Wenjie Wang, Yiyan Xu, Fuli Feng, Xinyu Lin, Xiangnan He, and Tat-Seng Chua.
\newblock Diffusion recommender model.
\newblock {\em arXiv preprint arXiv:2304.04971}, 2023.

\bibitem{wang2023conditional}
Yu~Wang, Zhiwei Liu, Liangwei Yang, and Philip~S Yu.
\newblock Conditional denoising diffusion for sequential recommendation.
\newblock {\em arXiv preprint arXiv:2304.11433}, 2023.

\bibitem{xu2019modeling}
Lei Xu, Maria Skoularidou, Alfredo Cuesta-Infante, and Kalyan Veeramachaneni.
\newblock Modeling tabular data using conditional gan.
\newblock {\em Advances in Neural Information Processing Systems}, 32, 2019.

\bibitem{camino2019improving}
Ramiro~D Camino, Christian~A Hammerschmidt, and Radu State.
\newblock Improving missing data imputation with deep generative models.
\newblock {\em arXiv preprint arXiv:1902.10666}, 2019.

\bibitem{chawla2002smote}
Nitesh~V Chawla, Kevin~W Bowyer, Lawrence~O Hall, and W~Philip Kegelmeyer.
\newblock Smote: synthetic minority over-sampling technique.
\newblock {\em Journal of artificial intelligence research}, 16:321--357, 2002.

\bibitem{ronneberger2015u}
Olaf Ronneberger, Philipp Fischer, and Thomas Brox.
\newblock U-net: Convolutional networks for biomedical image segmentation.
\newblock In {\em Medical Image Computing and Computer-Assisted
  Intervention--MICCAI 2015: 18th International Conference, Munich, Germany,
  October 5-9, 2015, Proceedings, Part III 18}, pages 234--241. Springer, 2015.

\bibitem{schroff2015facenet}
Florian Schroff, Dmitry Kalenichenko, and James Philbin.
\newblock Facenet: A unified embedding for face recognition and clustering.
\newblock In {\em Proceedings of the IEEE conference on computer vision and
  pattern recognition}, pages 815--823, 2015.

\bibitem{chen2022analog}
Ting Chen, Ruixiang Zhang, and Geoffrey Hinton.
\newblock Analog bits: Generating discrete data using diffusion models with
  self-conditioning.
\newblock {\em arXiv preprint arXiv:2208.04202}, 2022.

\bibitem{gorishniy2021revisiting}
Yury Gorishniy, Ivan Rubachev, Valentin Khrulkov, and Artem Babenko.
\newblock Revisiting deep learning models for tabular data.
\newblock {\em Advances in Neural Information Processing Systems},
  34:18932--18943, 2021.

\bibitem{ho1998use}
Siu~Lau Ho and Min Xie.
\newblock The use of arima models for reliability forecasting and analysis.
\newblock {\em Computers \& industrial engineering}, 35(1-2):213--216, 1998.

\bibitem{hewamalage2021recurrent}
Hansika Hewamalage, Christoph Bergmeir, and Kasun Bandara.
\newblock Recurrent neural networks for time series forecasting: Current status
  and future directions.
\newblock {\em International Journal of Forecasting}, 37(1):388--427, 2021.

\bibitem{zhou2021informer}
Haoyi Zhou, Shanghang Zhang, Jieqi Peng, Shuai Zhang, Jianxin Li, Hui Xiong,
  and Wancai Zhang.
\newblock Informer: Beyond efficient transformer for long sequence time-series
  forecasting.
\newblock In {\em Proceedings of the AAAI conference on artificial
  intelligence}, volume~35, pages 11106--11115, 2021.

\bibitem{cao2018brits}
Wei Cao, Dong Wang, Jian Li, Hao Zhou, Lei Li, and Yitan Li.
\newblock Brits: Bidirectional recurrent imputation for time series.
\newblock {\em Advances in neural information processing systems}, 31, 2018.

\bibitem{vaswani2017attention}
Ashish Vaswani, Noam Shazeer, Niki Parmar, Jakob Uszkoreit, Llion Jones,
  Aidan~N Gomez, {\L}ukasz Kaiser, and Illia Polosukhin.
\newblock Attention is all you need.
\newblock {\em Advances in neural information processing systems}, 30, 2017.

\bibitem{gu2021efficiently}
Albert Gu, Karan Goel, and Christopher R{\'e}.
\newblock Efficiently modeling long sequences with structured state spaces.
\newblock {\em arXiv preprint arXiv:2111.00396}, 2021.

\bibitem{kong2020diffwave}
Zhifeng Kong, Wei Ping, Jiaji Huang, Kexin Zhao, and Bryan Catanzaro.
\newblock Diffwave: A versatile diffusion model for audio synthesis.
\newblock {\em arXiv preprint arXiv:2009.09761}, 2020.

\bibitem{vahdat2020nvae}
Arash Vahdat and Jan Kautz.
\newblock Nvae: A deep hierarchical variational autoencoder.
\newblock {\em Advances in neural information processing systems},
  33:19667--19679, 2020.

\bibitem{saremi2019neural}
Saeed Saremi and Aapo Hyvarinen.
\newblock Neural empirical bayes.
\newblock {\em arXiv preprint arXiv:1903.02334}, 2019.

\bibitem{kim2018disentangling}
Hyunjik Kim and Andriy Mnih.
\newblock Disentangling by factorising.
\newblock In {\em International Conference on Machine Learning}, pages
  2649--2658. PMLR, 2018.

\bibitem{li2020behrt}
Yikuan Li, Shishir Rao, Jos{\'e} Roberto~Ayala Solares, Abdelaali Hassaine,
  Rema Ramakrishnan, Dexter Canoy, Yajie Zhu, Kazem Rahimi, and Gholamreza
  Salimi-Khorshidi.
\newblock Behrt: transformer for electronic health records.
\newblock {\em Scientific reports}, 10(1):1--12, 2020.

\bibitem{bhoi2021personalizing}
Suman Bhoi, Mong~Li Lee, Wynne Hsu, Hao Sen~Andrew Fang, and Ngiap~Chuan Tan.
\newblock Personalizing medication recommendation with a graph-based approach.
\newblock {\em ACM Transactions on Information Systems (TOIS)}, 40(3):1--23,
  2021.

\bibitem{poncette2019clinical}
Akira-Sebastian Poncette, Claudia Spies, Lina Mosch, Monique Schieler, Steffen
  Weber-Carstens, Henning Krampe, Felix Balzer, et~al.
\newblock Clinical requirements of future patient monitoring in the intensive
  care unit: qualitative study.
\newblock {\em JMIR medical informatics}, 7(2):e13064, 2019.

\bibitem{naik2022legal}
Nithesh Naik, BM~Hameed, Dasharathraj~K Shetty, Dishant Swain, Milap Shah,
  Rahul Paul, Kaivalya Aggarwal, Sufyan Ibrahim, Vathsala Patil, Komal Smriti,
  et~al.
\newblock Legal and ethical consideration in artificial intelligence in
  healthcare: who takes responsibility?
\newblock {\em Frontiers in surgery}, page 266, 2022.

\bibitem{baowaly2019synthesizing}
Mrinal~Kanti Baowaly, Chia-Ching Lin, Chao-Lin Liu, and Kuan-Ta Chen.
\newblock Synthesizing electronic health records using improved generative
  adversarial networks.
\newblock {\em Journal of the American Medical Informatics Association},
  26(3):228--241, 2019.

\bibitem{anderson1965iterative}
Donald~G Anderson.
\newblock Iterative procedures for nonlinear integral equations.
\newblock {\em Journal of the ACM (JACM)}, 12(4):547--560, 1965.

\bibitem{dhariwal2021diffusion}
Prafulla Dhariwal and Alexander Nichol.
\newblock Diffusion models beat gans on image synthesis.
\newblock {\em Advances in Neural Information Processing Systems},
  34:8780--8794, 2021.

\bibitem{karras2022elucidating}
Tero Karras, Miika Aittala, Timo Aila, and Samuli Laine.
\newblock Elucidating the design space of diffusion-based generative models.
\newblock {\em arXiv preprint arXiv:2206.00364}, 2022.

\bibitem{johnson2016mimic}
Alistair~EW Johnson, Tom~J Pollard, Lu~Shen, Li-wei~H Lehman, Mengling Feng,
  Mohammad Ghassemi, Benjamin Moody, Peter Szolovits, Leo Anthony~Celi, and
  Roger~G Mark.
\newblock Mimic-iii, a freely accessible critical care database.
\newblock {\em Scientific data}, 3(1):1--9, 2016.

\bibitem{pal2015evaluation}
Anita Pal, Ajeet~Kumar Gautam, and Yogendra~Narain Singh.
\newblock Evaluation of bioelectric signals for human recognition.
\newblock {\em Procedia Computer Science}, 48:746--752, 2015.

\bibitem{zimetbaum2003use}
Peter~J Zimetbaum and Mark~E Josephson.
\newblock Use of the electrocardiogram in acute myocardial infarction.
\newblock {\em New England Journal of Medicine}, 348(10):933--940, 2003.

\bibitem{chatterjee2020review}
Shubhojeet Chatterjee, Rini~Smita Thakur, Ram~Narayan Yadav, Lalita Gupta, and
  Deepak~Kumar Raghuvanshi.
\newblock Review of noise removal techniques in ecg signals.
\newblock {\em IET Signal Processing}, 14(9):569--590, 2020.

\bibitem{delaney2019synthesis}
Anne~Marie Delaney, Eoin Brophy, and Tomas~E Ward.
\newblock Synthesis of realistic ecg using generative adversarial networks.
\newblock {\em arXiv preprint arXiv:1909.09150}, 2019.

\bibitem{romero2021deepfilter}
Francisco~P Romero, David~C Pi{\~n}ol, and Carlos~R V{\'a}zquez-Seisdedos.
\newblock Deepfilter: An ecg baseline wander removal filter using deep learning
  techniques.
\newblock {\em Biomedical Signal Processing and Control}, 70:102992, 2021.

\bibitem{laguna1997database}
Pablo Laguna, Roger~G Mark, A~Goldberg, and George~B Moody.
\newblock A database for evaluation of algorithms for measurement of qt and
  other waveform intervals in the ecg.
\newblock In {\em Computers in cardiology 1997}, pages 673--676. IEEE, 1997.

\bibitem{moody1984noise}
George~B Moody, W~Muldrow, and Roger~G Mark.
\newblock A noise stress test for arrhythmia detectors.
\newblock {\em Computers in cardiology}, 11(3):381--384, 1984.

\bibitem{deng2019arcface}
Jiankang Deng, Jia Guo, Niannan Xue, and Stefanos Zafeiriou.
\newblock Arcface: Additive angular margin loss for deep face recognition.
\newblock In {\em Proceedings of the IEEE/CVF conference on computer vision and
  pattern recognition}, pages 4690--4699, 2019.

\bibitem{tangermann2012review}
Michael Tangermann, Klaus-Robert M{\"u}ller, Ad~Aertsen, Niels Birbaumer,
  Christoph Braun, Clemens Brunner, Robert Leeb, Carsten Mehring, Kai~J Miller,
  Gernot Mueller-Putz, et~al.
\newblock Review of the bci competition iv.
\newblock {\em Frontiers in neuroscience}, page~55, 2012.

\bibitem{zhang2021artificial}
Qian Zhang, Jie Lu, and Yaochu Jin.
\newblock Artificial intelligence in recommender systems.
\newblock {\em Complex \& Intelligent Systems}, 7:439--457, 2021.

\bibitem{he2016fast}
Xiangnan He, Hanwang Zhang, Min-Yen Kan, and Tat-Seng Chua.
\newblock Fast matrix factorization for online recommendation with implicit
  feedback.
\newblock In {\em Proceedings of the 39th International ACM SIGIR conference on
  Research and Development in Information Retrieval}, pages 549--558, 2016.

\bibitem{sun2019bert4rec}
Fei Sun, Jun Liu, Jian Wu, Changhua Pei, Xiao Lin, Wenwu Ou, and Peng Jiang.
\newblock Bert4rec: Sequential recommendation with bidirectional encoder
  representations from transformer.
\newblock In {\em Proceedings of the 28th ACM international conference on
  information and knowledge management}, pages 1441--1450, 2019.

\bibitem{liang2018variational}
Dawen Liang, Rahul~G Krishnan, Matthew~D Hoffman, and Tony Jebara.
\newblock Variational autoencoders for collaborative filtering.
\newblock In {\em Proceedings of the 2018 world wide web conference}, pages
  689--698, 2018.

\bibitem{ren2020sequential}
Ruiyang Ren, Zhaoyang Liu, Yaliang Li, Wayne~Xin Zhao, Hui Wang, Bolin Ding,
  and Ji-Rong Wen.
\newblock Sequential recommendation with self-attentive multi-adversarial
  network.
\newblock In {\em Proceedings of the 43rd international ACM SIGIR conference on
  research and development in information retrieval}, pages 89--98, 2020.

\bibitem{xiao2022tackling}
Zhisheng Xiao, Karsten Kreis, and Arash Vahdat.
\newblock Tackling the generative learning trilemma with denoising diffusion
  {GAN}s.
\newblock In {\em International Conference on Learning Representations}, 2022.

\bibitem{moraffah2021causal}
Raha Moraffah, Paras Sheth, Mansooreh Karami, Anchit Bhattacharya, Qianru Wang,
  Anique Tahir, Adrienne Raglin, and Huan Liu.
\newblock Causal inference for time series analysis: Problems, methods and
  evaluation.
\newblock {\em Knowledge and Information Systems}, 63:3041--3085, 2021.

\bibitem{meng2022interpretability}
Chuizheng Meng, Loc Trinh, Nan Xu, James Enouen, and Yan Liu.
\newblock Interpretability and fairness evaluation of deep learning models on
  mimic-iv dataset.
\newblock {\em Scientific Reports}, 12(1):7166, 2022.

\bibitem{muthukumar2021stochastic}
Pratyush Muthukumar and Jie Zhong.
\newblock A stochastic time series model for predicting financial trends using
  nlp.
\newblock {\em arXiv preprint arXiv:2102.01290}, 2021.

\bibitem{liu2018table}
Tianyu Liu, Kexiang Wang, Lei Sha, Baobao Chang, and Zhifang Sui.
\newblock Table-to-text generation by structure-aware seq2seq learning.
\newblock In {\em Proceedings of the AAAI conference on artificial
  intelligence}, volume~32, 2018.

\bibitem{niu2022fusion}
Ke~Niu, You Lu, Xueping Peng, and Jingni Zeng.
\newblock Fusion of sequential visits and medical ontology for mortality
  prediction.
\newblock {\em Journal of Biomedical Informatics}, 127:104012, 2022.

\end{thebibliography}

\newpage

\appendix
\appendixpage
\addappheadtotoc

\begin{appendices}

\section{Multinomial Diffusion for Categorical Data}
\label{appendix:multinomialdiffusion}

In this section, we introduce multinomial diffusion \cite{hoogeboom2021argmax}, which is designed to process categorical data. For $K$ categorical data, each variable is encoded as a one-hot vector, denoted as $ \mathbf{x}_t \in \{0, 1\} ^K $. Using categorical distribution, the multinomial diffusion process is formulated utilising the uniform diffusion noise schedule $ \beta_t $ and categorical distribution $ \mathcal{C} $:

\begin{equation} \label{eq:multinomial_1}
    q(\mathbf{x}_{t}|\mathbf{x}_{t-1})= \mathcal{C}(\mathbf{x}_t;(1-\beta_t)\mathbf{x}_{t-1} + \beta_t/K).
\end{equation}

Then, we can compute the probability of any $ \mathbf{x}_{t} $ given $ \mathbf{x}_{0} $ with $ \alpha_t := 1 - \beta_t $ and $ \bar{\alpha}_t := \prod^{t}_{\tau=0} \alpha_{\tau} $:

\begin{equation} \label{eq:multinomial_2}
    q(\mathbf{x}_{t}|\mathbf{x}_{0})= \mathcal{C}(\mathbf{x}_{t}; \bar{\alpha}_t \mathbf{x}_{0} + (1 - \bar{\alpha}_t)/K).
\end{equation}

According to Hoogeboom \textit{et al}. \cite{hoogeboom2021argmax}, the distribution for the preceding time step $ t-1 $ can be expressed from the value $ \mathbf{x}_t $ at the next step and the ground truth value $ \mathbf{x}_0 $ as following:

\begin{equation} \label{eq:multinomial_3}
    q(\mathbf{x}_{t-1}|\mathbf{x}_t, \mathbf{x}_0) = \mathcal{C}(\mathbf{x}_{t-1}; \tilde{\theta} / A)
\end{equation}

where $ \tilde{\theta}=[\alpha_t \mathbf{x}_t + (1-\alpha_t)/K]\odot[\bar{\alpha}_{t-1} \mathbf{x}_0 + (1-\bar{\alpha}_{t-1})/K] $ and $ A $ is a normalising constant that guarantees the cumulative total of all probabilities equals one. It is noteworthy that $ q(\mathbf{x}_{t-1}|\mathbf{x}_{t}, \mathbf{x}_0) $ simplifies to $ q(\mathbf{x}_{t-1}|\mathbf{x}_t) $, following the Markov property of the forward process.

Lastly, the reverse process $ p_\theta(\mathbf{x}_{t-1} | \mathbf{x}_{t}) $ is also learned via a deep neural network, denoted as $ q(\mathbf{x}_{t-1} | \mathbf{x}_t, \hat{\mathbf{x}}_{0}(\mathbf{x}_t, t)) $ where $ \hat{\mathbf{x}}_0 $ is the predicted probabilities by the deep neural network. The model is trained by minimising KL divergence (the $ L_{t-1} $ term in Equation \ref{eq:ddpm_kl}) between the true distribution and predicted one as:

\begin{equation} \label{eq:multinomial_4}
    D_\mathrm{KL}{(q(\mathbf{x}_{t-1}|\mathbf{x}_t,\mathbf{x}_0)} \hspace*{1mm} || \hspace*{1mm} {p_\theta(\mathbf{x}_{t-1}|\mathbf{x}_t))} = D_\mathrm{KL} (\mathcal{C} (\tilde{\theta}(\mathbf{x}_t, \mathbf{x}_0)) \hspace*{1mm} || \hspace*{1mm} \mathcal{C} (\tilde{\theta} (\mathbf{x}_t, \hat{\mathbf{x}}_0))).
\end{equation}

\newpage
\section{A Detailed Table on Generative Diffusion Models on Structured Data}
\label{appendix:comprehensive}

\begin{table}[h!]
\caption{A detailed table on generative diffusion models for structured data, including their frameworks, datasets used for experiments and accessible code links}
\centering
\scriptsize
\begin{tabular}{c|c|c|c|c|c}
\hline
\textbf{Paper} & \textbf{Year} & \textbf{Task} & \textbf{Framework} & \textbf{Dataset} & \textbf{Code} \\ \hline
TabDDPM \cite{kotelnikov2022tabddpm} & 2022 & Tabular Generation & DDPM & \begin{tabular}[c]{@{}c@{}} Abalone, Adult ROC, Buddy, California Housing, Cardio, \\ Churn Modelling, Diabetes, Facebook Comments Volume, \\ Gesture Phase, Higgs Small, House 16H, Insurance, King, \\ MiniBooNE, Wilt \end{tabular} & \href{https://github.com/rotot0/tab-ddpm}{Link} \\ \hline
SOS \cite{kim2022sos} & 2022 & Tabular Generation & SDE & \begin{tabular}[c]{@{}c@{}} Buddy, Default, Satimage, Shoppers, Surgical, WeatherAUS \end{tabular} & \href{https://github.com/JayoungKim408/SOS}{Link} \\ \hline
STaSy \cite{kim2022stasy} & 2023 & Tabular Generation & SDE & \begin{tabular}[c]{@{}c@{}} Bean, Beijing, Credit, Crowdsource, Contraceptive, Default, \\ HTRU, Magic, News, Obesity, Phishing, Robot, Shoppers, \\ Shuttle, Spambase \end{tabular} & \href{https://github.com/JayoungKim408/STaSy}{Link} \\ \hline
CoDi \cite{lee2023codi} & 2023 & Tabular Generation & DDPM & \begin{tabular}[c]{@{}c@{}} Absent, Bank, CMC, Customer, Drug, Faults, Heart, Insurance, \\ Obesity, Seismic, Stroke \end{tabular} & \href{https://github.com/ChaejeongLee/CoDi}{Link} \\ \hline
TabCSDI \cite{zheng2022diffusion} & 2022 & Tabular Imputation & DDPM & \begin{tabular}[c]{@{}c@{}} Breast, Census, Concrete, COVID-19, Diabetes, Libras, Wine \end{tabular} & \href{https://github.com/pfnet-research/TabCSDI}{Link} \\ \hline
TSGM \cite{lim2023regular} & 2023 & Time Series Generation & SDE & \begin{tabular}[c]{@{}c@{}} Air, AI4I, Energy, Occupancy, Stocks \end{tabular} & N/A \\ \hline
CSDI \cite{tashiro2021csdi} & 2021 & \begin{tabular}[c]{@{}c@{}}Time Series Imputation \\ Time Series Forecasting \end{tabular} & DDPM & \begin{tabular}[c]{@{}c@{}} Air Quality, PhysioNet 2012 Challenge \end{tabular} & \href{https://github.com/ermongroup/CSDI}{Link} \\ \hline
SSSD$^\mathrm{S4}$ \cite{alcaraz2022diffusion} & 2022 & \begin{tabular}[c]{@{}c@{}}Time Series Imputation \\ Time Series Forecasting \end{tabular} & DDPM & ETTm1, MuJoCo, PTB-XL, Solar & \href{https://github.com/AI4HealthUOL/SSSD}{Link} \\ \hline
DSPD/CSPD \cite{bilovs2022modeling} & 2022 & \begin{tabular}[c]{@{}c@{}}Time Series Imputation \\ Time Series Forecasting\end{tabular} & SDE & \begin{tabular}[c]{@{}c@{}} CIR, Lorenz, OU, Predator-prey, Sine, Sink  \end{tabular} & N/A \\ \hline
TimeGrad \cite{rasul2021autoregressive} & 2021 & Time Series Forecasting & DDPM & \begin{tabular}[c]{@{}c@{}} Electricity, Exchange, Solar, Taxi, Traffic, Wiki \end{tabular} & \href{https://github.com/zalandoresearch/pytorch-ts}{Link} \\ \hline
ScoreGrad \cite{yan2021scoregrad} & 2021 & Time Series Forecasting & SDE & \begin{tabular}[c]{@{}c@{}} Electricity, Exchange, Solar, Taxi, Traffic, Wiki \end{tabular} & \href{https://github.com/yantijin/ScoreGradPred}{Link} \\ \hline
D$^{3}$VAE \cite{li2022generative} & 2022 & Time Series Forecasting & DDPM & \begin{tabular}[c]{@{}c@{}} Electricity, ETTm1, ETTh1, Traffic, Weather, Wind \end{tabular} & \href{https://github.com/PaddlePaddle/PaddleSpatial/tree/main/research/D3VAE}{Link} \\ \hline
EHRDiff \cite{yuan2023ehrdiff} & 2023 & EHR Generation & SDE & MIMIC-III & \href{https://github.com/sczzz3/ehrdiff}{Link} \\ \hline
TabDDPM \cite{ceritli2023synthesizing} & 2023 & EHR Generation & DDPM & MIMIC-III & N/A \\ \hline
MedDiff \cite{he2023meddiff} & 2023 & EHR Generation & SDE & MIMIC-III, Patient Treatment Classification & N/A \\ \hline
EHR-DPM \cite{kuo2023synthetic} & 2023 & EHR Generation & DDPM & \begin{tabular}[c]{@{}c@{}}MIMIC-III, EuResist \end{tabular} & N/A \\ \hline
TDSTF \cite{chang2023tdstf} & 2023 & EHR Forecasting & DDPM & MIMIC-III & \href{https://github.com/PingChang818/TDSTF}{Link} \\ \hline
SSSD-ECG \cite{alcaraz2023diffusion} & 2022 & Biosignal Generation & DDPM & PTB-XL & \href{https://github.com/AI4HealthUOL/SSSD-ECG}{Link} \\ \hline
DeScoD-ECG \cite{li2023descod} & 2022 & Biosignal Enhancement & DDPM & \begin{tabular}[c]{@{}c@{}} MIT-BIH Noise Stress Test Database, QT Database \end{tabular} & \href{https://github.com/huayuliarizona/score-based-ecg-denoising}{Link} \\ \hline
DS-DDPM \cite{duan2023domain} & 2022 & Biosignal Enhancement & DDPM & \begin{tabular}[c]{@{}c@{}} BCI-Competition-IV dataset \end{tabular} & \href{https://github.com/duanyiqun/DS-DDPM}{Link} \\ \hline
CODIGEM \cite{walker2022recommendation} & 2022 & RecSys & DDPM & \begin{tabular}[c]{@{}c@{}} Amazon Electronics, MovieLens-1m, MovieLens-20m \end{tabular} & \href{https://github.com/WorldChanger01/CODIGEM}{Link} \\ \hline
DiffuRec \cite{li2023diffurec} & 2023 & RecSys & DDPM & \begin{tabular}[c]{@{}c@{}} Beauty, MovieLens-1M, Steam, Toys \end{tabular} & N/A \\ \hline
DiffRec (Du \textit{et al}.) \cite{du2023sequential} & 2023 & RecSys & DDPM & \begin{tabular}[c]{@{}c@{}} Beauty, MovieLens-1M, Toys \end{tabular} & N/A \\ \hline
DiffRec (Wang \textit{et al}.) \cite{wang2023diffusion} & 2023 & RecSys & DDPM & \begin{tabular}[c]{@{}c@{}} Amazon-book, MovieLens-1M, Yelp \end{tabular} & \href{https://github.com/YiyanXu/DiffRec}{Link} \\ \hline
CDDRec \cite{wang2023conditional} & 2023 & RecSys & DDPM & \begin{tabular}[c]{@{}c@{}} Beauty, Home, Office, Tools \end{tabular} & N/A \\ \hline
\end{tabular}
\label{table:comprehensivetable}
\end{table}

\end{appendices}

\end{document}